\documentclass{SCIS2026}
\usepackage{enumitem}
\usepackage{xcolor}
\usepackage{colortbl}
\usepackage[figuresright]{rotating}
\usepackage{afterpage}

\hypersetup{
    colorlinks=true,
    linkcolor=blue!70!black,
    citecolor=green!50!black,
    urlcolor=blue!70!black
}

\definecolor{hcolor}{HTML}{4472C4}
\definecolor{vcolor}{HTML}{ED7D31}
\definecolor{hvcolor}{HTML}{70AD47}
\definecolor{constraintc}{HTML}{5B9BD5}
\definecolor{scorec}{HTML}{FFC000}
\definecolor{continuousc}{HTML}{FF6384}
\definecolor{hybridc}{HTML}{A855F7}
\definecolor{headerblue}{HTML}{2E5090}
\definecolor{lightblue}{HTML}{D6E4F0}
\definecolor{lightgray}{HTML}{F2F2F2}

\newcommand{\ie}{\textit{i.e.}}
\newcommand{\eg}{\textit{e.g.}}
\newcommand{\etal}{\textit{et al.}}

\begin{document}
\ArticleType{REVIEW}
\Year{2026}
\Month{January}
\Vol{68}
\No{1}
\DOI{}
\ArtNo{}
\ReceiveDate{}
\ReviseDate{}
\AcceptDate{}
\OnlineDate{}
\AuthorMark{}
\AuthorCitation{}

\title{A Survey on Federated Causal Discovery and Inference}{Title for citation}

\author[1]{Xianjie GUO}{}
\author[2]{Yuwei WANG}{}
\author[2]{Guodu XIANG}{}
\author[3]{Xiaoli TANG}{}
\author[2]{\protect\\Kui YU}{{yukui@hfut.edu.cn}}
\author[3]{Han YU}{{han.yu@ntu.edu.sg}}
\author[4]{Qiang YANG}{}


\address[1]{School of Computer Science, Nanjing University of Posts and Telecommunications, Nanjing 210023, China}
\address[2]{School of Computer Science and Information Engineering, Hefei University of Technology, Hefei 230601, China}
\address[3]{College of Computing and Data Science, Nanyang Technological University, Singapore 639798}
\address[4]{Department of Data Science and Artificial Intelligence, The Hong Kong Polytechnic University, Hong Kong 999077, China}

\abstract{Causal reasoning, which encompasses the discovery of causal structures and the inference of causal effects, is fundamental to data-driven decision making.
In practice, data for reliable causal analysis are often distributed across institutions and cannot be centralized due to privacy regulations or communication constraints.
Federated learning (FL) addresses this by enabling collaborative analysis without raw data sharing, giving rise to the rapidly growing field of \textbf{federated causal discovery (FCD) and inference (FCI)}.
However, the interdisciplinary nature of this field and the absence of a comprehensive survey present barriers to entry for researchers. This paper bridges that gap by providing a systematic review through multi-dimensional taxonomies.
Grounded in the three core design decisions underlying any FCD solution, namely how structures are learned, how data are partitioned, and what structural knowledge each party obtains, we organize \textbf{FCD} along three axes: \emph{methodological paradigm} (constraint-based, score-based, continuous-optimization, and hybrid), \emph{federation topology} (horizontal, vertical, and hybrid), and \emph{structural scope} (global versus local).
We further examine key practical dimensions, including temporal dynamics, data heterogeneity, missing data, and non-identical variable sets.
For \textbf{FCI}, we categorize methods by \emph{target estimand} (average versus individualized/conditional treatment effects) and by \emph{estimation strategy}, from classical weighting methods to modern deep generative architectures.
Unlike prior works that treat FCD and FCI separately, we formalize their connection as complementary stages of a unified federated causal reasoning pipeline, where FCD supplies the structural knowledge required for valid effect estimation in FCI.
Finally, we highlight their shared concerns regarding privacy, communication efficiency, theoretical guarantees, and application domains, and conclude by identifying open challenges for future research.
}

\keywords{Causal Discovery, Causal Inference, Federated Learning, Bayesian Networks, Causal Structure Learning, Treatment Effect Estimation, Privacy-Preserving Learning}

\maketitle


\section{Introduction}\label{sec:intro}

Causal reasoning lies at the heart of scientific inquiry: understanding \emph{why} things happen enables prediction under intervention and robust decision making.
Two complementary tasks form the backbone of computational causal analysis.
\textbf{Causal discovery} aims to recover the directed acyclic graph (DAG) that encodes cause-effect relationships among a set of variables from observational (and sometimes interventional) data~\cite{spirtes2000causation,chickering2002optimal,zheng2018dags}.
In contrast, \textbf{causal inference} seeks to quantify the magnitude of causal effects, typically treatment effects, from data that may be confounded by non-random treatment assignment~\cite{pearl2009causality,imbens2015causal}.

In many real-world settings, the data needed for these tasks resides across multiple institutions, such as hospitals, banks, IoT devices, or research centers, that are unable or unwilling to share raw records.
Privacy legislation (e.g., GDPR, HIPAA), commercial sensitivity, and sheer data volume all conspire to prevent centralization.
\textbf{Federated learning} (FL)~\cite{mcmahan2017communication,kairouz2021advances} offers a principled solution: participants collaboratively train a shared model by exchanging only intermediate statistics or model parameters, while keeping raw data local.

\begin{figure}[t!]
	\centering
	\includegraphics[width=1.0\linewidth]{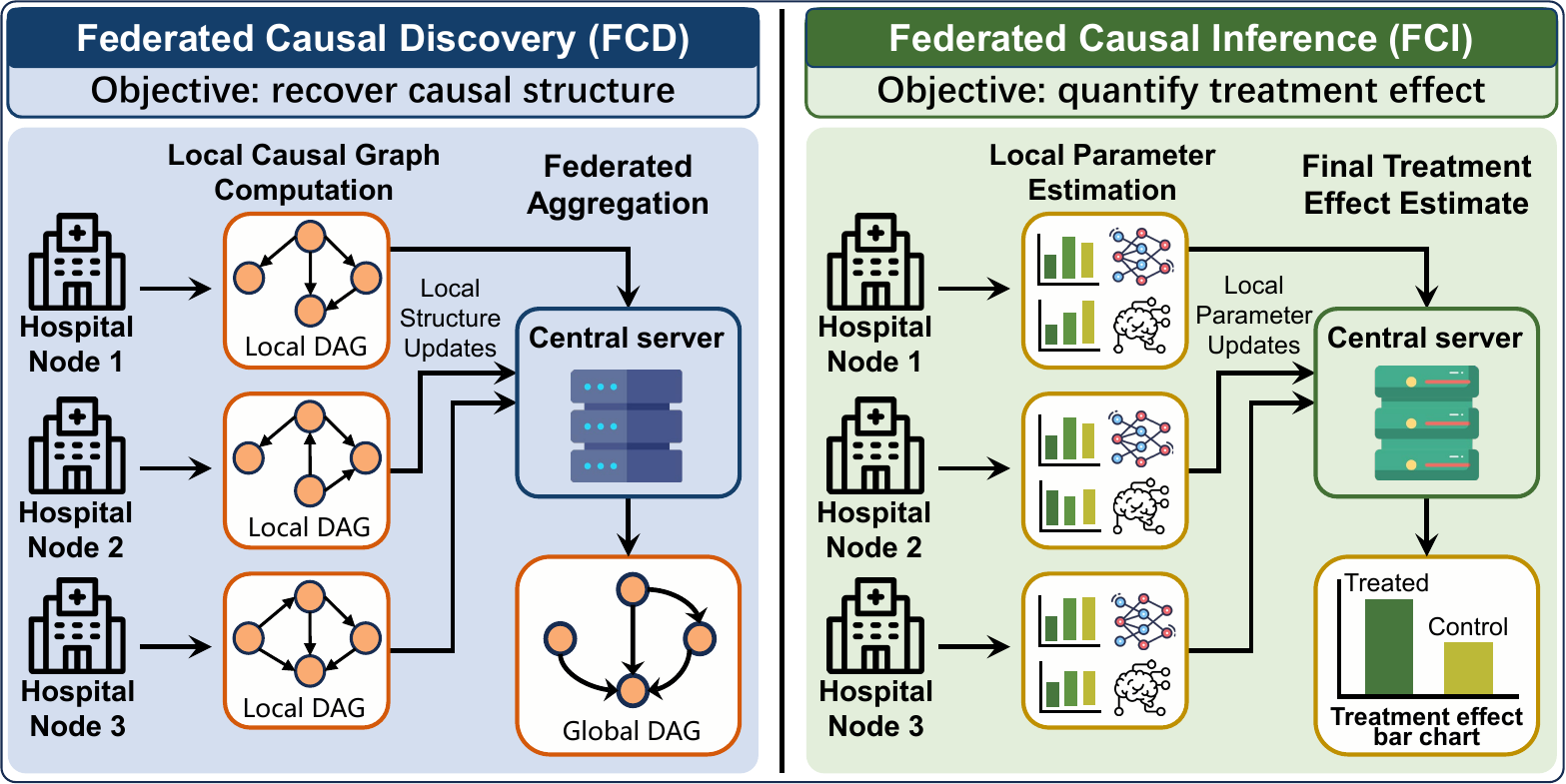}
	\caption{An intuitive illustration of federated causal learning frameworks. (a) \textbf{Federated Causal Discovery (FCD)}: Distributed clients collaboratively recover the global causal graph topology (\ie, skeleton and edge orientations) via decentralized statistics or gradients without sharing raw data. (b) \textbf{Federated Causal Inference (FCI)}: Distributed clients collaboratively quantify the specific causal effect (\eg, Average Treatment Effect, ATE) of a treatment variable on an outcome by jointly adjusting for confounders and aligning propensity scores across sites.} 
	\label{fig:fcd_fci_intuitive}
\end{figure}

When bridging causal frameworks with FL, the focus shifts based on the specific analytical objective, resulting in two entirely new paradigms as illustrated in Figure~\ref{fig:fcd_fci_intuitive}:
\begin{itemize}[leftmargin=*,nosep]
    \item \textbf{Federated Causal Discovery (FCD)} focuses on ``recovering the structure''. As shown in Figure~\ref{fig:fcd_fci_intuitive}(a), instead of pooling raw data to find cause-effect links, individual clients compute local structural properties (such as conditional independencies or graph gradients) and securely aggregate them at a central server to piece together a unified global directed acyclic graph (DAG).
    \item \textbf{Federated Causal Inference (FCI)} focuses on ``quantifying the effect''. As shown in Figure~\ref{fig:fcd_fci_intuitive}(b), assuming a target treatment and outcome are specified, clients collaboratively model treatment assignments and outcomes across shifting populations to accurately estimate a global effect metric (\eg, the Average Treatment Effect, ATE) without exposing private patient covariates.
\end{itemize}

While FL has achieved remarkable success in supervised prediction tasks (image classification, next-word prediction, and clinical risk scoring), its integration with causal methods introduces distinctive challenges:
\begin{enumerate}[leftmargin=*,nosep]
    \item \textbf{Structural constraints.} Unlike regression, causal discovery must produce a DAG satisfying acyclicity, which is a global, combinatorial constraint that is non-trivial to enforce in a distributed setting.
    \item \textbf{Heterogeneity in causal mechanisms and distributions.} A defining challenge across federated causal tasks is that clients do not merely differ in marginal data distributions, as label-shift or covariate-shift models in standard FL assume, but exhibit heterogeneous \emph{causal mechanisms}: the conditional distributions $P(X_i \mid \mathrm{Pa}(X_i))$ governing local data-generating processes vary across sites, and clients may even observe different sets of variables. This structural heterogeneity manifests in two interrelated ways: in federated causal discovery, it means that no single causal graph faithfully represents all clients; in federated causal inference, it means that treatment assignment mechanisms and covariate distributions may differ drastically across sites, invalidating naive pooling or meta-analysis.
    \item \textbf{Identifiability.} Causal conclusions are sensitive to the sufficiency of conditioning sets, the availability of interventional data, and functional-form assumptions. Distributing these requirements across clients raises new questions about identifiability.
\end{enumerate}

The field has grown rapidly: from early work on privacy-preserving Bayesian network structure learning in the mid-2000s~\cite{wright2004privacy,gou2007learning} to recent methods published at venues.
Despite this momentum, no existing survey provides a unified, systematic treatment of both federated causal discovery and federated causal inference with fine-grained methodological taxonomies. Rocchi~\etal~\cite{rocchi2025federated} offer a medical-domain perspective on federated causal discovery, but do not cover causal inference or provide a fine-grained methodological taxonomy. Several federated learning surveys~\cite{kairouz2021advances,li2021survey} touch upon privacy-preserving analytics but do not specifically address causal reasoning. Most recently, Deng~\etal~\cite{deng2026survey} present a broad survey of the interplay between causality and federated learning, covering both how FL enables decentralized causal analysis and how causality enhances FL systems in dimensions such as interpretability, generalizability, and adversarial robustness.
Although their work provides a valuable high-level overview of this emerging landscape, our survey differs fundamentally in \emph{scope}, \emph{taxonomy}, and \emph{depth}:
\begin{enumerate}[leftmargin=*,nosep]
    \item \textbf{Focused scope.} We concentrate exclusively on the core algorithmic challenges of federated causal discovery and federated causal inference, rather than adopting the broader ``Causality for FL'' perspective (\eg, using causality to improve FL robustness or fairness).
    \item \textbf{Multi-dimensional taxonomies.} We propose the first three-axis taxonomy for FCD (methodology $\times$ federation topology $\times$ structural scope) and a two-axis taxonomy for FCI (estimand $\times$ estimation strategy), enabling principled, fine-grained comparison of existing methods along dimensions not previously synthesized.
    \item \textbf{Technical depth.} We provide detailed algorithmic analyses, formal problem formulations, privacy-theoretic discussions, and systematic comparisons of theoretical guarantees across methods.
\end{enumerate}
In addition, we are the first to explicitly formalize and analyze the connection between FCD and FCI as complementary stages of a unified federated causal reasoning pipeline.

\begin{figure}[t!]
	\centering
	\includegraphics[width=1.0\linewidth]{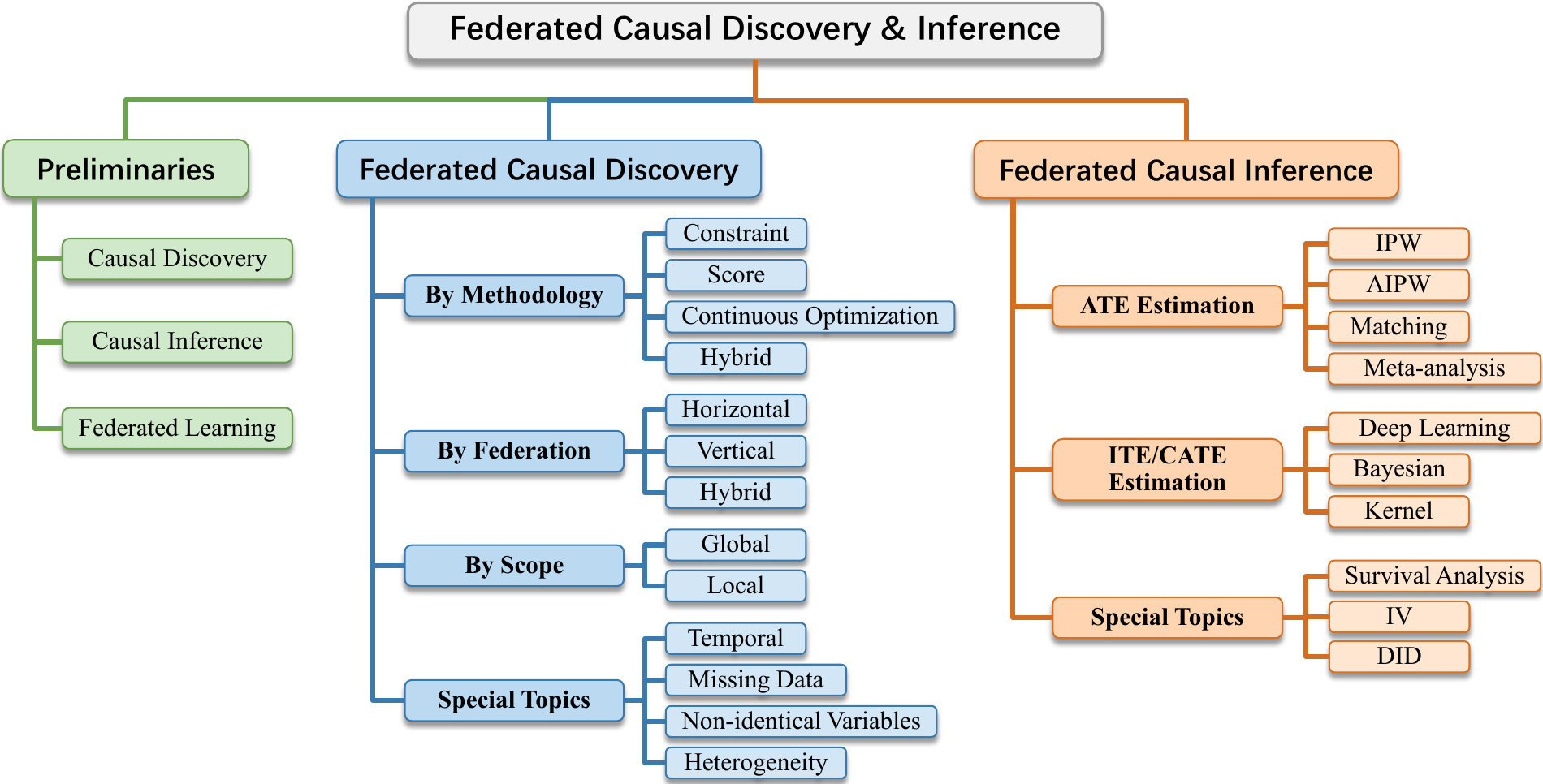}
	\caption{Organization and scope of this survey.}
	\label{fig:survey_scope}
\end{figure}

The main contributions of this survey are:
\begin{enumerate}[leftmargin=*,nosep]
    \item We present a comprehensive, focused survey of federated causal discovery and inference within a unified framework. Unlike the broader ``Causal-FL'' perspective in~\cite{deng2026survey}, which spans both ``FL for Causality'' and ``Causality for FL'' (including interpretability, generalizability, and robustness enhancement), our work provides an in-depth algorithmic and theoretical treatment dedicated to the core problems of distributed structure learning and effect estimation. To our knowledge, this level of dedicated, fine-grained methodological analysis for both FCD and FCI has not been previously offered.
    \item We propose novel multi-dimensional taxonomies, namely a three-axis framework for FCD (by methodology, federation topology, and structural scope) and a two-axis framework for FCI (by estimand and estimation strategy), that enable principled, granular comparison of existing work beyond single-dimensional categorizations.
    \item We provide detailed summary tables, systematic comparison along privacy mechanisms, communication efficiency, theoretical guarantees, and practical considerations, together with a curated bibliography to serve as a reference for researchers and practitioners.
    \item We explicitly formalize and analyze the connection between FCD and FCI as complementary stages of a unified federated causal reasoning pipeline, and identify open problems and promising directions for future research.
\end{enumerate}

\noindent\textbf{Organization.} 
The overall structure of this survey is depicted in Figure~\ref{fig:survey_scope}. 
Following the preliminaries in Section~\ref{sec:prelim}, we systematically survey the methodologies of federated causal discovery (Section~\ref{sec:fcd}) and federated causal inference (Section~\ref{sec:fci}). 
Section~\ref{sec:connection} elaborates on the intrinsic connections between these two tasks. 
Furthermore, Section~\ref{sec:apps} highlights practical applications, while Section~\ref{sec:future} outlines potential future avenues. 
Section~\ref{sec:conclusion} concludes this work.

\section{Preliminaries}\label{sec:prelim}
 
\subsection{Causal Discovery}\label{sec:prelim_cd}
 
Causal discovery aims to learn a DAG $\mathcal{G} = (\mathbf{V}, \mathbf{E})$ over a set of variables $\mathbf{V} = \{X_1, \dots, X_d\}$ from data.  The edges in $\mathbf{E}$ encode direct causal relationships: $X_i \to X_j$ means $X_i$ is a direct cause of $X_j$.
Under a structural equation model (SEM), each variable is generated as:
\begin{equation}
    X_j = f_j(\mathrm{Pa}(X_j), \epsilon_j), \quad j = 1, \dots, d,
    \label{eq:sem}
\end{equation}
where $\mathrm{Pa}(X_j)$ denotes the parent set of $X_j$ in $\mathcal{G}$, $f_j$ is a (possibly nonlinear) function, and $\epsilon_j$ are jointly independent noise variables.
 
\textbf{Assumptions and identifiability.}
Recovering $\mathcal{G}$ from data is possible only under a set of standard assumptions that link the graph $\mathcal{G}$ to the observed joint distribution $P$.
The \emph{causal Markov condition} requires that each variable be independent of its non-descendants given its parents, so that $\mathcal{G}$ entails a collection of conditional independencies in $P$.
The \emph{faithfulness} assumption imposes the converse, requiring that every conditional independence present in $P$ is entailed by $\mathcal{G}$ through d-separation, thereby ruling out independencies that arise from the accidental cancellation of causal effects.
Taken together, the Markov and faithfulness assumptions yield the equivalence that $X_i \perp\!\!\!\perp X_j \mid \mathbf{S}$ holds in $P$ if and only if $\mathbf{S}$ d-separates $X_i$ and $X_j$ in $\mathcal{G}$~\cite{spirtes2000causation,pearl2009causality}, which is precisely what allows structural information to be inferred from observed statistical (in)dependencies.
Two further assumptions are routinely adopted: \emph{causal sufficiency}, namely that all common causes of the measured variables are themselves measured so that there are no latent confounders, and \emph{acyclicity}, which restricts the hypothesis space to DAGs.
Even when all of these conditions hold, observational data alone identify $\mathcal{G}$ only up to its \emph{Markov equivalence class} (MEC), the set of DAGs encoding the same conditional independencies, which is represented by a completed partially directed acyclic graph (CPDAG)~\cite{spirtes2000causation,chickering2002optimal}.
Resolving the remaining orientation ambiguity requires additional leverage, such as restrictions on the functional form $f_j$ or the noise distribution $\epsilon_j$ in the SEM of Eq.~\eqref{eq:sem}~\cite{shimizu2006linear,shimizu2011directlingam}, or access to interventional data~\cite{hauser2012characterization,eberhardt2005number}.
 
Building on these assumptions, classical approaches to causal discovery fall into three broad paradigms, with a fourth category addressing more specialized settings.
 
\textbf{Constraint-based methods} (e.g., PC~\cite{spirtes2000causation}, FCI~\cite{colombo2012learning,zhang2008completeness}) test conditional independence (CI) relations to first learn a skeleton and then orient edges via orientation rules.  They are assumption-light but sensitive to the reliability of CI tests, especially with limited samples.  The PC algorithm operates in a layer-wise fashion: at depth $\ell$, it tests $X_i \perp\!\!\!\perp X_j \mid \mathbf{S}$ for all conditioning sets $\mathbf{S}$ of size $\ell$ drawn from the adjacency set, removing edges upon conditional independence~\cite{spirtes2000causation,colombo2014order}.
 
\textbf{Score-based methods} (e.g., GES~\cite{chickering2002optimal}, K2~\cite{cooper1992bayesian}) search the space of DAGs (or equivalence classes) by optimizing a scoring criterion such as BIC:
\begin{equation}
    \text{BIC}(\mathcal{G}; \mathcal{D}) = -2 \sum_{j=1}^{d} \hat{\ell}_j(\mathrm{Pa}_{\mathcal{G}}(X_j); \mathcal{D}) + |\mathcal{G}| \cdot \log n,
    \label{eq:bic}
\end{equation}
where $\hat{\ell}_j$ is the maximized local log-likelihood for variable $X_j$ given its parents, $|\mathcal{G}|$ is the number of free parameters, and $n$ is the sample size.  The search is typically greedy or uses heuristics to navigate the super-exponential DAG space~\cite{chickering2002optimal,heckerman1995learning}.
 
\textbf{Continuous optimization methods} reformulate the combinatorial DAG learning problem as a continuous program.  The seminal NOTEARS~\cite{zheng2018dags} formulation introduces a smooth acyclicity constraint $h(\mathbf{W}) = \mathrm{tr}(e^{\mathbf{W} \circ \mathbf{W}}) - d = 0$ on the weighted adjacency matrix $\mathbf{W}$, enabling gradient-based optimization.  Extensions handle nonlinear models~\cite{zheng2020learning,lachapelle2020gradient}, non-Gaussian noise~\cite{shimizu2006linear,shimizu2011directlingam}, and time-series data~\cite{pamfil2020dynotears,tank2021neural}.
 
\textbf{Hybrid methods} combine constraint-based and score-based ideas, using CI tests to prune the search space before a score-based procedure refines the structure.  Representative examples include MMHC~\cite{tsamardinos2006max}, which builds a skeleton via the MMPC test and then runs greedy hill-climbing over it, and ARGES~\cite{nandy2018high}, which confines the GES search to a conditional-independence graph while providing high-dimensional consistency guarantees.
 
\subsection{Causal Inference}\label{sec:prelim_ci}
 
Causal inference focuses on estimating the effect of a treatment $A$ on an outcome $Y$ in the presence of confounders $\mathbf{X}$.  Under the potential outcomes framework~\cite{splawa1990application}, each unit $i$ has two potential outcomes, $Y_i(1)$ and $Y_i(0)$, where $Y_i(a)$ denotes the outcome that would be observed under treatment $A=a$.  The \emph{individual treatment effect} (ITE) is defined as the unit-level contrast $\tau_i = Y_i(1) - Y_i(0)$.  Since at most one potential outcome is observed for any unit, the ITE is not identifiable in general, so attention turns to its population average.  The \emph{average treatment effect} (ATE) is the expectation of the ITE over the population, $\tau = \mathbb{E}[Y(1) - Y(0)]$.
 
Key identification assumptions include: (i)~\emph{unconfoundedness} (ignorability): $Y(0), Y(1) \perp\!\!\!\perp A \mid \mathbf{X}$; (ii)~\emph{positivity} (overlap): $0 < P(A=1 \mid \mathbf{X}=\mathbf{x}) < 1$ for all $\mathbf{x}$ in the support; and (iii)~\emph{consistency}: $Y = Y(A)$~\cite{imbens2015causal}.  Under these, the ATE is identifiable from observational data via the adjustment formula:
\begin{equation}
    \tau = \mathbb{E}_{\mathbf{X}}\left[\mathbb{E}[Y \mid A=1, \mathbf{X}] - \mathbb{E}[Y \mid A=0, \mathbf{X}]\right].
    \label{eq:adjustment}
\end{equation}
 
Standard estimators include:
\textbf{Inverse Probability Weighting} (IPW)~\cite{horvitz1952generalization}, which reweights observations by the inverse of the propensity score $e(\mathbf{X}) = P(A=1 \mid \mathbf{X})$:
\begin{equation}
    \hat{\tau}_{\text{IPW}} = \frac{1}{n}\sum_{i=1}^{n}\left[\frac{A_i Y_i}{\hat{e}(\mathbf{X}_i)} - \frac{(1-A_i)Y_i}{1-\hat{e}(\mathbf{X}_i)}\right];
    \label{eq:ipw}
\end{equation}
\textbf{Outcome Regression} (G-computation)~\cite{robins1986new}, which models $\mu_a(\mathbf{X}) = \mathbb{E}[Y \mid A=a, \mathbf{X}]$ directly;
\textbf{Augmented IPW} (AIPW)~\cite{bang2005doubly}, which combines both and enjoys double robustness, remaining consistent if \emph{either} the propensity or outcome model is correctly specified:
\begin{equation}
    \hat{\tau}_{\text{AIPW}} = \frac{1}{n}\sum_{i=1}^{n}\left[\hat{\mu}_1(\mathbf{X}_i) - \hat{\mu}_0(\mathbf{X}_i) + \frac{A_i(Y_i - \hat{\mu}_1(\mathbf{X}_i))}{\hat{e}(\mathbf{X}_i)} - \frac{(1-A_i)(Y_i - \hat{\mu}_0(\mathbf{X}_i))}{1 - \hat{e}(\mathbf{X}_i)}\right];
    \label{eq:aipw}
\end{equation}
and \textbf{Matching} methods~\cite{stuart2010matching}, which pair treated and control units with similar covariates.
The \emph{conditional average treatment effect} (CATE), $\tau(\mathbf{x}) = \mathbb{E}[Y(1) - Y(0) \mid \mathbf{X}=\mathbf{x}]$, captures heterogeneity and is a central target in personalized decision making~\cite{kunzel2019metalearners,kennedy2023towards}.
 
\subsection{Federated Learning}\label{sec:prelim_fl}
 
\begin{figure}[b!]
	\centering
	\includegraphics[width=1.0\linewidth]{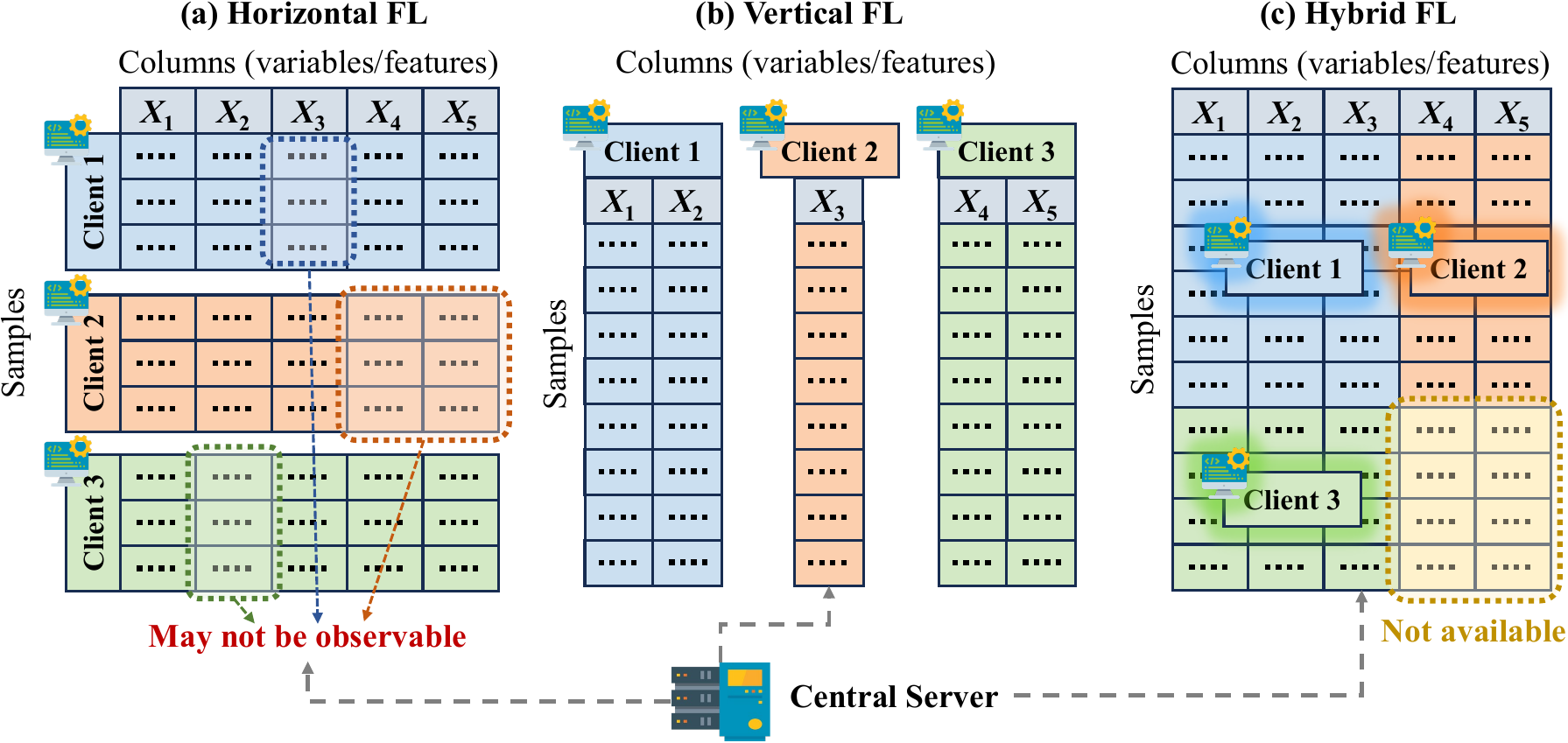}
	\caption{Three federated data partitioning paradigms: (a)~horizontal, (b)~vertical, and (c)~hybrid.}
	\label{fig:fl_partition}
\end{figure}
 
Federated learning enables $K$ clients, each holding a private dataset $\mathcal{D}_k$, to collaboratively train a model without sharing raw data~\cite{mcmahan2017communication,kairouz2021advances}.  A central server coordinates the process by aggregating local updates. As illustrated in Figure~\ref{fig:fl_partition}, federated learning settings can be categorized by how data is partitioned across clients.
The canonical federated optimization objective is:
\begin{equation}
    \min_{\boldsymbol{\theta}} \; F(\boldsymbol{\theta}) = \sum_{k=1}^{K} \frac{n_k}{n} F_k(\boldsymbol{\theta}), \quad \text{where} \quad F_k(\boldsymbol{\theta}) = \frac{1}{n_k}\sum_{i=1}^{n_k} \mathcal{L}(\boldsymbol{\theta}; \mathbf{z}_i^{(k)}),
    \label{eq:fl_objective}
\end{equation}
with $n = \sum_{k=1}^K n_k$ denoting the total sample size, $\mathcal{L}$ the loss function, and $\mathbf{z}_i^{(k)}$ the $i$-th data point at client $k$.
 
\textbf{Horizontal FL} (sample partitioning): clients share the same feature space but hold different samples.  The canonical algorithm FedAvg~\cite{mcmahan2017communication} averages locally trained model parameters: $\boldsymbol{\theta}^{(t+1)} = \sum_{k=1}^K \frac{n_k}{n} \boldsymbol{\theta}_k^{(t+1)}$.  Variants such as FedProx~\cite{li2020federated_fedprox}, SCAFFOLD~\cite{karimireddy2020scaffold}, and FedNova~\cite{wang2020tackling} address challenges arising from non-IID data and heterogeneous local computation.
 
\textbf{Vertical FL} (feature partitioning): clients observe different features of the same set of entities, linked by a shared identifier~\cite{yang2019federated,liu2024vertical}.  Formally, $\mathbf{X}^{(k)} \in \mathbb{R}^{n \times d_k}$ with $\sum_k d_k = d$ and shared row indices.
 
\textbf{Hybrid FL}: both sample and feature spaces differ across clients, representing the most general and challenging setting~\cite{yang2019federated}.
 
Key challenges in FL include statistical heterogeneity (non-IID data)~\cite{sattler2019robust,li2022federated}, communication efficiency~\cite{wang2022communication,wu2022communication}, and privacy.  Privacy-enhancing techniques include differential privacy (DP)~\cite{dwork2014algorithmic,abadi2016deep}, secure multi-party computation (SMPC)~\cite{knott2021crypten,bonawitz2017practical}, and homomorphic encryption~\cite{gentry2009fully,acar2018survey}.

\section{Federated Causal Discovery}\label{sec:fcd}

Federated causal discovery (FCD) aims to learn a causal structure from data distributed across multiple clients without centralizing raw data.
We organize the literature along three complementary taxonomic dimensions, as presented in Table~\ref{tab:fcd_methods}: \emph{methodological paradigm} (\S\ref{sec:fcd_method}), \emph{federation topology} (\S\ref{sec:fcd_topo}), and \emph{structural scope} (\S\ref{sec:fcd_scope}).
To our knowledge, this three-axis taxonomic framework, jointly considering methodological paradigm, federation topology, and structural scope, provides the most fine-grained and systematic organization of federated causal discovery methods to date, enabling a more nuanced comparison than the single-dimensional categorizations used in prior discussions~\cite{deng2026survey,rocchi2025federated}.
We then discuss important cross-cutting themes (\S\ref{sec:fcd_special}).

\afterpage{
\clearpage                        
\begin{sidewaystable}[p]   
\setlength{\tabcolsep}{13.6pt} 
\centering
\caption{Comprehensive comparison of federated causal discovery methods. \textbf{Method paradigm}: CB = Constraint-Based, SB = Score-Based, CO = Continuous Optimization, HY = Hybrid. \textbf{Federation}: H = Horizontal, V = Vertical, H+V = Both. \textbf{Scope}: G = Global, L = Local. \textbf{Privacy}: DP = Differential Privacy, SMPC = Secure Multi-Party Computation, SS = Summary Statistics only. \textbf{Heterogeneity}: Homo. = Homogeneous data assumed, Hetero. = Heterogeneous data supported. \textbf{Missing}: whether the method explicitly handles missing data ($\checkmark$ = supported). \textbf{Time}: whether the method supports temporal/time-series causal discovery ($\checkmark$ = supported). \textbf{Non-id.\ Vars}: whether the method handles non-identical (overlapping but different) variable sets across clients ($\checkmark$ = supported).}
\label{tab:fcd_methods}
\footnotesize
\renewcommand{\arraystretch}{1.15}
\begin{tabular}{@{}llcccccccc@{}}
\toprule
\rowcolor{headerblue}
\textcolor{white}{\textbf{Method}} & \textcolor{white}{\textbf{Venue}} & \textcolor{white}{\textbf{Paradigm}} & \textcolor{white}{\textbf{Fed.}} & \textcolor{white}{\textbf{Scope}} & \textcolor{white}{\textbf{Privacy}} & \textcolor{white}{\textbf{Hetero.}} & \textcolor{white}{\textbf{Missing}} & \textcolor{white}{\textbf{Time}} & \textcolor{white}{\textbf{Non-id. Vars}} \\
\midrule
DARLS~\cite{ye2024federated}            & TPAMI'24   & \cellcolor{scorec!30}SB   & \cellcolor{hcolor!20}H   & G & SS   & Hetero. &       &       & \\
P-TPDA~\cite{gou2007learning}            & SNPD'07   & \cellcolor{constraintc!20}CB  & \cellcolor{hcolor!20}H   & G & SS    & Homo. &       &       & \\
PP-BN(H)~\cite{samet2009privacy}         & ICCSE'09  & \cellcolor{scorec!30}SB  & \cellcolor{hcolor!20}H   & G & SMPC  & Homo. &       &       & \\
PP-BN(V)~\cite{wright2004privacy}        & KDD'04    & \cellcolor{scorec!30}SB  & \cellcolor{vcolor!20}V   & G & SMPC  & --    &       &       & \\
DBNSL~\cite{na2010distributed}           & ISIE'10   & \cellcolor{scorec!30}SB  & \cellcolor{hcolor!20}H   & G & SS    & Homo. &       &       & \\
BNSL-DD~\cite{chen2003learning}          & ICDM'03    & \cellcolor{hybridc!20}HY  & \cellcolor{vcolor!20}V   & G & SS    & --    &       &       & \\
NOTEARS-ADMM~\cite{ng2022towards}        & AISTATS'22& \cellcolor{continuousc!20}CO  & \cellcolor{hcolor!20}H   & G & SS    & Homo. &       &       & \\
RFCD~\cite{mian2022regret}               & KDD-W'22  & \cellcolor{scorec!30}SB  & \cellcolor{hcolor!20}H   & G & SS    & Homo. &       &       & \\
FedDAG~\cite{gao2023feddag}              & TMLR'23   & \cellcolor{continuousc!20}CO  & \cellcolor{hcolor!20}H   & G & SS    & Hetero.&      &       & \\
FedPC~\cite{huang2023towards}            & TBD'23    & \cellcolor{constraintc!20}CB  & \cellcolor{hcolor!20}H   & G & SS    & Homo. &       &       & \\
PERI~\cite{mian2023nothing}              & AISTATS'23& \cellcolor{scorec!30}SB  & \cellcolor{hcolor!20}H   & G & DP    & Homo. &       &       & \\
FedC$^2$SL~\cite{wang2023towards}        & ECML-PKDD'23   & \cellcolor{constraintc!20}CB  & \cellcolor{hcolor!20}H   & G & SS    & Homo. &       &       & \\
FedCDH~\cite{li2024federated}            & ICLR'24   & \cellcolor{constraintc!20}CB  & \cellcolor{hcolor!20}H   & G & SS    & Hetero.&      &       & \\
FedCausal~\cite{yang2024federated}       & AAAI'24   & \cellcolor{continuousc!20}CO  & \cellcolor{hcolor!20}H   & G & SS    & Hetero.&      &       & \\
FedCSL~\cite{guo2024fedcsl}              & AAAI'24   & \cellcolor{constraintc!20}CB  & \cellcolor{hcolor!20}H   & G & SS    & Homo. &       &       & \\
FedACD~\cite{guo2024sample}              & IJCAI'24  & \cellcolor{continuousc!20}CO  & \cellcolor{hcolor!20}H   & G & SS    & Hetero.&      &       & \\
FedCASL~\cite{liu2024federated}          & CYBER'24  & \cellcolor{continuousc!20}CO  & \cellcolor{hcolor!20}H   & G & SS    & Homo. &       &       & \\
CCaT~\cite{tregubov2024tool}             & SCBM'24   & \cellcolor{hybridc!20}HY  & \cellcolor{hcolor!20}H   & G & SS    & Hetero.&      &       & \\
Bloom~\cite{qiu2024interventional}       & TNSE'25   & \cellcolor{continuousc!20}CO  & \cellcolor{hcolor!20}H   & G & SS    & Homo. &       &       & \\
VertiBayes~\cite{van2024vertibayes}      & CIS'24    & \cellcolor{scorec!30}SB  & \cellcolor{vcolor!20}V   & G & SMPC  & --    & \checkmark &  & \\
FedECD~\cite{guo2024enhancing}           & NeurIPS-W'24& \cellcolor{constraintc!20}CB & \cellcolor{hcolor!20}H  & G & SS    & Homo. &       &       & \\
FedLCS~\cite{yu2025federated}            & SCIS'25   & \cellcolor{constraintc!20}CB  & \cellcolor{hcolor!20}H   & L & SS    & Homo. &       &       & \\
eFedLCS~\cite{rong2025efficient}         & IJCRS'25  & \cellcolor{constraintc!20}CB  & \cellcolor{hcolor!20}H   & L & SS    & Homo. &       &       & \\
FedCSL-HD~\cite{ding2025federated}       & NNICE'25  & \cellcolor{constraintc!20}CB  & \cellcolor{hcolor!20}H   & G & SS    & Homo. &       &       & \\
FedImpCSL~\cite{shi2025federated}        & KBS'25    & \cellcolor{constraintc!20}CB  & \cellcolor{hcolor!20}H   & G & SS    & Homo. & \checkmark &  & \\
FDBNL/PFDBNL~\cite{chen2025federated}    & TAI'25    & \cellcolor{continuousc!20}CO  & \cellcolor{hcolor!20}H   & G & SS    & Hetero.&      & \checkmark & \\
FedGCL~\cite{mohanty2025federated}       & ICLR'25   & \cellcolor{hybridc!20}HY  & \cellcolor{vcolor!20}V   & G & SS    & Hetero.&      & \checkmark & \\
FedPuzzle~\cite{li2025fedpuzzle}         & INS'25    & \cellcolor{constraintc!20}CB  & \cellcolor{hcolor!20}H   & G & SS    & Hetero.&      &       & \checkmark \\
FedCDnv~\cite{wang2025federated}         & ICML'25   & \cellcolor{constraintc!20}CB  & \cellcolor{hcolor!20}H   & G & SS    & Homo. &       &       & \checkmark \\
FMT-NBN~\cite{yang2025federated}         & IISE-T'25   & \cellcolor{continuousc!20}CO  & \cellcolor{hcolor!20}H   & G & SS    & Hetero.&      &       & \checkmark \\
Fed-HC-aIPW~\cite{zanga2025federated}    & JBI'25    & \cellcolor{scorec!30}SB  & \cellcolor{hcolor!20}H   & G & SS    & Hetero.& \checkmark &  & \\
NOTEARS-PFL~\cite{liu2025federated}      & JBI'25    & \cellcolor{continuousc!20}CO  & \cellcolor{hcolor!20}H   & G & SS    & Hetero.&      &       & \\
I-PERI~\cite{baldo2026regret}            & ICML'26  & \cellcolor{scorec!30}SB  & \cellcolor{hcolor!20}H   & G & DP    & Hetero.&      &       & \\
FedISHC~\cite{chen2026horizontal}        & AAAI'26  & \cellcolor{hybridc!20}HY  & \cellcolor{hvcolor!20}H+V & G & SS    & Homo. &       &       & \\
Fed-Sparse-BNSL~\cite{fehri2026differentially}  & AISTATS'26 & \cellcolor{continuousc!20}CO  & \cellcolor{hcolor!20}H   & G & DP    & Hetero.&      &       & \\
FedGES~\cite{torrijos2026fedges}         & ML'26     & \cellcolor{scorec!30}SB  & \cellcolor{hcolor!20}H   & G & SS    & Hetero.&      &       & \\
fedCI-IOD~\cite{hahn2026federated}       & arXiv'26  & \cellcolor{constraintc!20}CB  & \cellcolor{hvcolor!20}H+V & G & SS    & Hetero.& \checkmark &  & \checkmark \\
FedGC~\cite{mohanty2026uncertainty}      & arXiv'26  & \cellcolor{hybridc!20}HY  & \cellcolor{vcolor!20}V   & G & SS    & Hetero.&      & \checkmark & \\
FedCDI~\cite{abyaneh2022federated}       & arXiv'22  & \cellcolor{hybridc!20}HY  & \cellcolor{hcolor!20}H   & G & SS    & Homo. &       &       & \\
FedECE~\cite{zhao2025fedece}            & TAI'25    & \cellcolor{constraintc!20}CB  & \cellcolor{hcolor!20}H   & G & SS    & Homo. &       &       & \\
\bottomrule
\end{tabular}
\end{sidewaystable}
}

\subsection{Taxonomy by Methodological Paradigm}\label{sec:fcd_method}

\subsubsection{Constraint-Based Methods}\label{sec:fcd_constraint}

Constraint-based FCD methods adapt the skeleton-discovery-then-orientation pipeline of classical algorithms (PC, FCI) to federated settings.
The core technical challenge is to perform \emph{federated conditional independence (CI) testing} without centralizing data.

\textbf{FedPC}~\cite{huang2023towards} is the first method to systematically adapt the PC algorithm to horizontal FL.
It introduces a \emph{layer-wise aggregation strategy}: at each conditioning depth, clients independently compute local CI statistics and share them with the server, which aggregates them into global CI decisions.
The \texttt{FedSkele} subroutine handles skeleton learning, while \texttt{FedOrien} resolves the \emph{consistent separation set} problem to accurately orient v-structures without centralizing data.
The layer-wise design ensures natural convergence without additional hyperparameters.

\textbf{FedC$^2$SL}~\cite{wang2023towards} proposes a federated CI test protocol where each client computes local test statistics that are securely aggregated into an unbiased estimate of the global statistic.
This enables any constraint-based algorithm to operate on the resulting global CI relations, and the authors demonstrate this with a federated adaptation achieving strong empirical performance.

\textbf{FedCDH}~\cite{li2024federated} tackles the particularly challenging setting of heterogeneous data distributions across clients.
It introduces a \emph{surrogate variable} indexed by client identity to model distribution shifts, and develops two key primitives: a \emph{federated CI test} (FCIT) using random feature approximations of kernel matrices, and a \emph{federated independent change principle} (FICP) that exploits causal asymmetry across heterogeneous domains.
Both are non-parametric, making no functional-form assumptions.

\textbf{fedCI-IOD}~\cite{hahn2026federated} addresses the most complex setting to date: non-identical variable sets, mixed data types, site-specific effects, and latent confounding.
It provides the first federated CI testing framework (fedCI) with a privacy-preserving client-server architecture, paired with a federated implementation of the IOD algorithm that supports both meta-analytic p-value aggregation and direct integration with federated CI tests, retaining all theoretical guarantees.

\textbf{FedCSL}~\cite{guo2024fedcsl} addresses scalability and accuracy simultaneously.
It introduces a \emph{local-to-global learning strategy} consisting of federated causal neighbor learning, global skeleton construction, and skeleton orientation, enabling scaling to high-dimensional data.
To handle uneven sample allocation across clients, it devises a novel weight-inference strategy that determines relative sample sizes without encryption, enabling weighted aggregation.

\textbf{FedCSL-HD}~\cite{ding2025federated} further tackles high-dimensional settings via a divide-and-conquer strategy: it builds \emph{federated unified clusters} to partition the variable space consistently across clients, learns cluster-level skeletons, and aggregates them through boundary-node-based fusion with a conflict-edge repair mechanism.

Several early works laid the groundwork.  \textbf{P-TPDA}~\cite{gou2007learning} learns local BN structures independently and combines them through CI testing.
\textbf{DBNSL}~\cite{na2010distributed} proposes aggregation based on frequently occurring substructures.
\textbf{PP-BN(H)}~\cite{samet2009privacy} uses secure multi-party computation protocols (secure exponentiation, factorial, and comparison) within the K2 algorithm.

\textbf{Limitations.}
Constraint-based methods inherit the well-known fragility of conditional independence (CI) testing: a single erroneous CI decision can propagate through the skeleton-discovery and orientation phases, producing cascading structural errors~\cite{spirtes2000causation}.
This fragility is amplified in the federated setting, where each global CI decision is an aggregate of noisy local statistics and where small local sample sizes render high-order CI tests (those conditioning on large sets) particularly unreliable.
Like their centralized counterparts, these methods typically rest on the faithfulness assumption and recover the structure only up to its Markov equivalence class, leaving a CPDAG with unoriented edges.
Most also assume causal sufficiency and statistical homogeneity across clients; FedCDH~\cite{li2024federated} and fedCI-IOD~\cite{hahn2026federated} are notable exceptions, but relax these assumptions at the cost of greater statistical or computational complexity (\eg, kernel-based federated CI tests introduce additional hyperparameters and heavier computation).
Finally, the layer-wise aggregation common to FedPC~\cite{huang2023towards} and FedCSL~\cite{guo2024fedcsl} incurs one communication round per conditioning depth, so communication cost grows with graph density and conditioning-set size.

\subsubsection{Score-Based Methods}\label{sec:fcd_score}

Score-based FCD methods adapt classical score-and-search paradigms to distributed settings, where the key challenge is evaluating and optimizing a global score function without data centralization.

\textbf{DARLS}~\cite{ye2024federated} is a theoretically grounded approach that searches over the space of \emph{orderings} $\pi$ using simulated annealing.
For each candidate ordering, a regularized log-likelihood score is evaluated via distributed optimization over DAGs compatible with $\pi$.
Crucially, since every DAG admits at least one topological ordering, searching over orderings inherently satisfies the acyclicity constraint.
The authors prove that the federated estimate converges to the oracle (centralized) solution at rate $O(\log(n)/\sqrt{m})$, where $n$ is the total sample size and $m$ is the minimum local sample size -- the first such guarantee for federated causal discovery.

\textbf{RFCD}~\cite{mian2022regret} introduces a novel privacy-preserving framework based on \emph{regret}.
Each client first discovers its best local model, then, given a server-proposed candidate DAG, reports only the regret -- the gap between its local score under the candidate and its best local score.
The server minimizes worst-case regret to find a global DAG.  Clients share neither data, parameters, nor local structures.
\textbf{PERI}~\cite{mian2023nothing} extends this by proving consistency of the regret-based score and employing the Laplace mechanism for $\epsilon$-differential privacy.
\textbf{I-PERI}~\cite{baldo2026regret} further extends the regret framework to \emph{unknown interventions} at the client level, introducing the $\Phi$-Markov equivalence class and a two-phase approach that first learns the CPDAG and then refines it by exploiting structural differences across clients.

\textbf{FedGES}~\cite{torrijos2026fedges} adapts the classical Greedy Equivalence Search (GES) to horizontal FL.
Clients learn local BNs, which are iteratively fused into a global structure.
The method preserves key theoretical properties of GES while introducing novel convergence criteria and fusion mechanisms to handle client heterogeneity.
Privacy is maintained by sharing only structural information.

\textbf{Fed-HC-aIPW}~\cite{zanga2025federated} is among the first to address \emph{missing data} in federated causal discovery.
It proposes a score-based algorithm that handles different missingness mechanisms across sources, evaluates the interaction between aggregation techniques and scoring criteria, and assesses violations of global missingness assumptions.

\textbf{PP-BN(V)}~\cite{wright2004privacy} is an early vertical FL approach that uses secret sharing to privately compute an approximation of the K2 score function across two parties holding different features.

\textbf{Limitations.}
Score-based methods must search a DAG (or ordering) space that grows super-exponentially with the number of variables, so scalability to high-dimensional problems remains their principal bottleneck, and greedy score-and-search procedures such as FedGES~\cite{torrijos2026fedges} can settle into suboptimal structures in the finite-sample regime, with their fusion heuristics offering no global-optimality guarantee.
Their accuracy hinges on the chosen score, which generally encodes parametric or distributional assumptions (\eg, linear-Gaussian or multinomial likelihoods), making them vulnerable to model misspecification, and like constraint-based methods, they typically identify the structure only up to its Markov equivalence class.
The federated guarantees also come with caveats: DARLS~\cite{ye2024federated} relies on computationally intensive simulated annealing, and its $O(\log(n)/\sqrt{m})$ rate is governed by the \emph{minimum} local sample size $m$, so a single small or low-quality client can dominate the error.
Regret-based methods~\cite{mian2022regret,mian2023nothing,baldo2026regret}, while attractive for privacy, require an expensive server-side combinatorial optimization and can be conservative; PERI, in particular, further trades structural accuracy for protection through differential-privacy noise.
Finally, iterative local-model fusion entails multiple communication rounds to reach consensus.

\subsubsection{Continuous Optimization Methods}\label{sec:fcd_continuous}

Continuous optimization FCD methods build on the NOTEARS paradigm, reformulating DAG learning as a continuous program amenable to gradient-based distributed optimization.

\textbf{NOTEARS-ADMM}~\cite{ng2022towards} is the seminal work in this category.
It applies the Alternating Direction Method of Multipliers (ADMM) to decompose the NOTEARS objective across clients, requiring only the exchange of model parameters during optimization.
The approach is demonstrated for both linear and nonlinear structural equation models (by employing multilayer perceptrons (MLP)).

\textbf{FedDAG}~\cite{gao2023feddag} proposes a two-level local model architecture separating \emph{graph structure learning} (GSL) from \emph{mechanism approximation} (MA).
During federated training, only the GSL components are shared, while MA parts are updated locally to accommodate data heterogeneity -- a design insight that neatly addresses the unique challenge of federated structural learning compared to standard FL.

\textbf{FedCausal}~\cite{yang2024federated} develops a global optimization process on the server that replaces traditional weighted averaging.
This design naturally enforces sparsity and acyclicity of the global graph.
The local and global optimization form a unified, \emph{explainable adaptive} objective that reduces to the centralized objective under statistical homogeneity, and gracefully handles heterogeneity.

\textbf{FedCASL}~\cite{liu2024federated} formulates the problem as a bi-level optimization on top of NOTEARS with a carefully designed sparse penalty to remove spurious edges under acyclicity constraints.

\textbf{FedACD}~\cite{guo2024sample} is the first to address \emph{sample quality heterogeneity}: the insight that different variable subspaces may have different data quality across clients.
It adaptively selects causal relationships learned under ``good'' variable spaces at each client and masks those from ``bad'' spaces, ensuring only high-quality learning results are communicated.

\textbf{Bloom}~\cite{qiu2024interventional} integrates both observational and interventional data through bilevel polynomial optimization with SDP relaxations, offering convergence and optimality guarantees, and extends to distributed settings.

\textbf{Fed-Sparse-BNSL}~\cite{fehri2026differentially} combines differential privacy with a proximal greedy coordinate descent that transmits only sparse edge updates per iteration rather than dense $d \times d$ matrices, drastically reducing both communication cost and privacy budget consumption.

\textbf{NOTEARS-PFL}~\cite{liu2025federated} introduces a \emph{personalized} federated BN learning method using sparse group LASSO to capture both shared structure and site-specific edges, motivated by multi-site neuroimaging analysis of major depressive disorder.

\textbf{FMT-NBN}~\cite{yang2025federated} proposes a federated multi-task framework for clients with \emph{overlapping and distinct variables}, using a nonparametric additive structural causal model with sparse group LASSO and a two-step FL framework that learns client similarities on the server.

\textbf{Limitations.}
As continuations of the NOTEARS paradigm, these methods cast DAG learning as a non-convex program and therefore converge only to stationary points, with no guarantee of recovering the global optimum.
The smooth acyclicity penalty (based on a matrix exponential or trace) costs $O(d^3)$ per evaluation, and the gradient-based formulation has a documented sensitivity to the scaling and standardization of variables, a vulnerability that is especially delicate in federated settings where preprocessing may differ across clients.
Most variants further commit to a specific functional form (linear SEMs or MLP-based mechanisms), limiting robustness under misspecification.
Communication is the dominant practical cost: exchanging $d \times d$ adjacency matrices or their gradients incurs $O(d^2)$ per round, and ADMM-style schemes such as NOTEARS-ADMM~\cite{ng2022towards} require many rounds to converge; the sparse-update scheme of Fed-Sparse-BNSL~\cite{fehri2026differentially} is a notable exception that mitigates this overhead.

\subsubsection{Hybrid and Other Methods}\label{sec:fcd_hybrid}

Several methods combine multiple paradigms or introduce entirely novel mechanisms.

\textbf{FedCDI}~\cite{abyaneh2022federated} uniquely targets \emph{interventional} data in federated settings.
Clients use neural causal discovery to form local beliefs about causal structure, which are aggregated on the server with quality-weighted contributions: clients with higher-quality interventional samples exert greater influence.

\textbf{CCaT}~\cite{tregubov2024tool} takes a collaborative, human-in-the-loop approach: sites evaluate and refine a shared causal model using private datasets, sharing only summary statistics or suggested causal relations while maintaining distinct local models.

\textbf{FedISHC}~\cite{chen2026horizontal} exploits higher-order cumulants for both horizontal \emph{and} vertical federated settings.
In linear non-Gaussian models, higher-order cumulants capture information about absent variables from the existing ones within each client, enabling identifiability in vertical FL without variable completeness.

\textbf{FedImpCSL}~\cite{shi2025federated} and \textbf{Fed-HC-aIPW}~\cite{zanga2025federated} both address missing data.
FedImpCSL proposes a local-to-global imputation strategy (FedLocalImp and FedGlobalImp) and a Federated Contribution Assessment Method (FedCAM) that jointly considers sample size and missing data rate for weighted aggregation.

\textbf{Limitations.}
Because this group combines disparate mechanisms, its members trade generality for specialized assumptions.
FedCDI~\cite{abyaneh2022federated} presupposes access to interventional data, which is costly or simply unavailable in many federated applications; CCaT~\cite{tregubov2024tool} relies on human-in-the-loop refinement, limiting automation and scalability; and FedISHC~\cite{chen2026horizontal} depends on higher-order cumulants that require large samples for stable estimation together with a linear non-Gaussian model assumption.
Imputation-based approaches such as FedImpCSL~\cite{shi2025federated} are only as reliable as their imputations and may introduce bias when missingness is non-ignorable.
As a heterogeneous collection, these methods also lack the shared theoretical framework and cross-method comparability that the more established paradigms enjoy.


\subsection{Taxonomy by Federation Topology}\label{sec:fcd_topo}

\subsubsection{Horizontal Federated Causal Discovery}

The vast majority of FCD methods operate under horizontal partitioning, where clients share the same variable set but hold different samples.
This setting is natural when institutions collect the same types of measurements (\eg, hospitals recording the same clinical variables) but serve different populations.
The main technical challenge is aggregating local structural information (CI test results, scores, or gradient updates) into a consistent global DAG.

Under statistical homogeneity (all clients share the same data-generating process), methods like FedPC~\cite{huang2023towards} and NOTEARS-ADMM~\cite{ng2022towards} can leverage the increased total sample size to improve over single-site discovery.
Under heterogeneity, methods like FedDAG~\cite{gao2023feddag}, FedCDH~\cite{li2024federated}, and FedCausal~\cite{yang2024federated} must carefully disentangle shared causal structure from client-specific mechanisms.

\subsubsection{Vertical Federated Causal Discovery}

In vertical FL, each client observes a different subset of variables for the same set of entities.
This creates a fundamentally different challenge: no single client can assess relationships between its variables and those of other clients without some form of cross-client communication.

PP-BN(V)~\cite{wright2004privacy} and BNSL-DD~\cite{chen2003learning} are early exemplars, using secret sharing and selective data transmission for cross-site structure learning.
VertiBayes~\cite{van2024vertibayes} handles an arbitrary number of parties and missing values.
FedGCL~\cite{mohanty2025federated} applies vertical FL to Granger causality in time-series settings, where clients share low-dimensional state representations rather than raw high-dimensional measurements.

\subsubsection{Hybrid Federated Causal Discovery}

Some recent methods address the more realistic setting where both sample and feature spaces differ across clients.

FedISHC~\cite{chen2026horizontal} provides a unified framework for both horizontal and vertical settings using higher-order cumulants, which are robust to absent variables.
fedCI-IOD~\cite{hahn2026federated} handles non-identical variable sets with mixed data types and site-specific effects, supporting partial horizontal and vertical overlap.

\subsection{Taxonomy by Structural Scope}\label{sec:fcd_scope}

\subsubsection{Global Causal Structure Learning}

Most FCD methods target the \emph{global} causal structure, namely the full DAG over all variables, which is the default and most studied setting. The computational complexity of learning a global DAG grows super-exponentially with the number of variables, and the federated communication overhead compounds this challenge.

\subsubsection{Local Causal Structure Learning}

\textbf{Local causal structure learning} (LCS) focuses on identifying the direct causes (parents) and effects (children) of a specific target variable, without learning the entire DAG.
This is particularly valuable in high-dimensional settings and for targeted applications such as finding risk factors for a specific disease.

\textbf{FedLCS}~\cite{yu2025federated} is the first federated LCS method, comprising three subroutines: federated local skeleton learning (\texttt{FLSke}), federated local skeleton orientation (\texttt{FLSori}), and federated local extension-and-backtracking orientation (\texttt{FLEori}).
\textbf{eFedLCS}~\cite{rong2025efficient} accelerates FedLCS through GPU-parallel CI testing and separation set optimization, addressing the computational bottleneck of repeated skeleton learning.

\subsection{Cross-Cutting Themes}\label{sec:fcd_special}

\subsubsection{Temporal and Dynamic Causal Discovery}

Time-series data presents additional challenges: the causal graph may have temporal edges, and Granger causality provides an alternative framework.

\textbf{FDBNL/PFDBNL}~\cite{chen2025federated} are the first methods for federated Dynamic Bayesian Network (DBN) learning from time-series data.
FDBNL handles homogeneous settings, whereas PFDBNL incorporates a proximal operator for heterogeneous time-series and provides the first convergence analysis for continuous-optimization-based structure learning in FL.

\textbf{FedGCL}~\cite{mohanty2025federated} proposes federated Granger causality learning for interdependent clients using state-space representations, enabling causal analysis of high-dimensional time-series without aggregating raw data.
\textbf{FedGC}~\cite{mohanty2026uncertainty} provides a systematic taxonomy of uncertainty sources (aleatoric vs.\ epistemic) in federated Granger causality and derives closed-form steady-state variance expressions.

\subsubsection{Data Heterogeneity}

Data heterogeneity is a critical challenge across all FCD paradigms, as clients may have different sample sizes, data quality, or even different underlying distributions.

Methods addressing heterogeneity include: FedDAG's separated GSL/MA architecture~\cite{gao2023feddag}, FedCDH's surrogate variable approach~\cite{li2024federated}, FedCausal's explainable adaptive optimization~\cite{yang2024federated}, FedACD's variable-space-level quality assessment~\cite{guo2024sample}, and NOTEARS-PFL's sparse group LASSO for shared vs.\ site-specific structures~\cite{liu2025federated}.

\subsubsection{Non-Identical Variable Sets}

In practice, clients often observe overlapping but non-identical variable sets, creating both \emph{absolute} latent variables (unobserved by all) and \emph{relative} latent variables (observed by some but not others).

\textbf{FedCDnv}~\cite{wang2025federated} develops theories to detect whether relationships involving non-overlapping variable pairs are definitively non-causal, and introduces a two-level priority selection strategy that identifies ``correct'' and ``good'' relationships.
\textbf{FedPuzzle}~\cite{li2025fedpuzzle} decomposes local causal graphs into \emph{star graph pieces} for reliable local discovery, then aggregates through edge-probability and star-graph weighting.
\textbf{FMT-NBN}~\cite{yang2025federated} treats this as a multi-task learning problem, fitting related but distinct DAGs for each client.

\subsubsection{Missing Data}

Missing data compounds the challenges of federated causal discovery, as missingness mechanisms may differ across clients.

\textbf{FedImpCSL}~\cite{shi2025federated} proposes local-to-global imputation with federated contribution assessment considering both sample size and missing data rate.
\textbf{Fed-HC-aIPW}~\cite{zanga2025federated} handles different missingness mechanisms across sources in a clinical endometrial cancer study.
\textbf{VertiBayes}~\cite{van2024vertibayes} is the first vertical FL method to handle missing values with arbitrary numbers of parties.

\subsubsection{Limited Samples and Bootstrapping}

\textbf{FedECD}~\cite{guo2024enhancing} addresses the performance degradation when each client holds very limited samples by employing a two-layer bootstrapping aggregation strategy at both client and server levels, improving skeleton and structure estimation under data scarcity.

\section{Federated Causal Inference}\label{sec:fci}

Federated causal inference (FCI) aims to estimate causal effects, typically the average treatment effect (ATE) or individualized/conditional average treatment effects (ITE/CATE), from data distributed across multiple sites.
Unlike FCD, which produces a structural graph, FCI quantifies the effect between treatment and outcome.

Although individual federated causal inference methods have appeared at leading venues, and Deng~\etal~provide a high-level discussion of federated causal inference within a broader Causal-FL framework~\cite{deng2026survey}, no prior work has offered a dedicated, fine-grained survey that systematically categorizes the federated causal inference literature by estimand type and estimation strategy with detailed algorithmic analysis. This section fills that gap by organizing existing methods along the dimensions of target estimand and estimation strategy.

We organize FCI methods by \emph{target estimand} and \emph{estimation strategy}, as presented in Table~\ref{tab:fci_methods}.

\afterpage{
\clearpage                        
\begin{sidewaystable}[p]  
\setlength{\tabcolsep}{13.6pt} 
\centering
\caption{Comprehensive comparison of federated causal inference methods. \textbf{Estimand}: ATE, CATE/ITE, ATT/ATC, DATE, HTE. \textbf{Strategy}: IPW, AIPW, Match., Bayes., DML, DL = Deep Learning, Meta = Meta-analysis variant, IV = Instrumental Variable, DR = Doubly Robust, CB/IDA = Constraint-Based structure learning followed by IDA-based effect estimation, M-est. = M-estimator, G-formula = Plug-in G-Formula.  \textbf{Comm.}: number of communication rounds (1-shot means a single round). \textbf{Privacy}: SS = Summary Statistics only, DP = Differential Privacy, Param.\ = Model Parameters shared, Repr.\ = Dimensionality-reduced Representations shared. \textbf{Outcome}: the type of outcome variable supported (Continuous, Binary, Binary/Cont., Time-to-event).}
\label{tab:fci_methods}
\footnotesize
\renewcommand{\arraystretch}{1.15}
\begin{tabular}{@{}llccccccc@{}}
\toprule
\rowcolor{headerblue}
\textcolor{white}{\textbf{Method}} & \textcolor{white}{\textbf{Venue}} & \textcolor{white}{\textbf{Estimand}} & \textcolor{white}{\textbf{Strategy}} & \textcolor{white}{\textbf{Fed. Type}} & \textcolor{white}{\textbf{Hetero.}} & \textcolor{white}{\textbf{Comm.}} & \textcolor{white}{\textbf{Privacy}} & \textcolor{white}{\textbf{Outcome}} \\
\midrule
FedCI~\cite{vo2022bayesian}               & UAI'22    & \cellcolor{continuousc!20}ATE/ITE  & Bayes./GP  & \cellcolor{hcolor!20}H & \checkmark & Iterative & SS    & Continuous \\
CausalRFF~\cite{vo2022adaptive}           & NeurIPS'22& \cellcolor{hybridc!20}ATE/CATE & Kernel     & \cellcolor{hcolor!20}H & \checkmark & Iterative & SS    & Continuous \\
FedMLE~\cite{xiong2023federated}      & SiM'23    & \cellcolor{scorec!30}ATE      & IPW-MLE    & \cellcolor{hcolor!20}H & \checkmark & 1-shot    & SS    & Binary/Cont. \\
FedIPW-MLE~\cite{xiong2023federated}  & SiM'23    & \cellcolor{scorec!30}ATE      & IPW-MLE    & \cellcolor{hcolor!20}H & \checkmark & 1-shot    & SS    & Binary/Cont. \\
FedAIPW~\cite{xiong2023federated}     & SiM'23    & \cellcolor{scorec!30}ATE      & AIPW       & \cellcolor{hcolor!20}H & \checkmark & 1-shot    & SS    & Binary/Cont. \\
MR\_L1~\cite{han2023multiply}             & NeurIPS'23& \cellcolor{scorec!30}ATE      & DR         & \cellcolor{hcolor!20}H & \checkmark & 1-shot    & SS    & Continuous \\
FedTEDVAE~\cite{almodovar2023federated}   & NeurIPS-W'23 & \cellcolor{constraintc!20}ITE   & DL/VAE     & \cellcolor{hcolor!20}H & \checkmark & Iterative & Param.& Continuous \\
FPSM~\cite{liu2025federated_p}            & AOR'25  & \cellcolor{scorec!30}ATE      & Match.     & \cellcolor{hcolor!20}H & --       & Iterative & SS    & Binary/Cont. \\
CLB-AIPW~\cite{guo2024collaborative}      & ICML'24   & \cellcolor{scorec!30}ATE      & IPW/DR     & \cellcolor{hcolor!20}H & \checkmark & 1-shot    & SS    & Continuous \\
COLA~\cite{hu2024collaborative}           & SiM'24    & \cellcolor{scorec!15}ATE/DATE & IPW        & \cellcolor{hcolor!20}H & --         & 2--4 rounds & SS  & Binary/Cont. \\
DC-QE~\cite{kawamata2024collaborative}    & ESWA'24   & \cellcolor{scorec!30}ATE      & Match./IPW & \cellcolor{hvcolor!20}H+V & --       & 1-shot    & Repr. & Continuous \\
FedTransTEE~\cite{makhija2024federated}   & arXiv'24   & \cellcolor{constraintc!20}ITE      & DL/Trans.  & \cellcolor{hcolor!20}H & \checkmark & Iterative & Param.& Continuous \\
Fed-IPTW~\cite{yin2026federated}               & TIST'26    & \cellcolor{constraintc!20}ITE      & IPW/DL     & \cellcolor{hcolor!20}H & \checkmark & Iterative & Param.& Continuous \\
FACE~\cite{han2025federated}              & JASA'25   & \cellcolor{scorec!30}ATE      & DR         & \cellcolor{hcolor!20}H & \checkmark & 1-shot    & SS    & Binary/Cont. \\
RIFL~\cite{guo2025robust}                 & JASA'25   & \cellcolor{scorec!30}ATE      & Meta/DR    & \cellcolor{hcolor!20}H & \checkmark & 1-shot    & SS    & Continuous \\
FDML~\cite{kang2025federated}             & Biometrics'25 & \cellcolor{scorec!30}ATE  & DML        & \cellcolor{hcolor!20}H & \checkmark & 1-shot    & SS    & Continuous \\
CausalFI~\cite{vo2025federated}           & ML'25     & \cellcolor{continuousc!20}ATE/ITE  & Bayes.     & \cellcolor{hcolor!20}H & \checkmark & Iterative & SS    & Continuous \\
FedECA~\cite{ogier2025fedeca}             & NC'25     & \cellcolor{scorec!30}ATE      & IPW/Cox    & \cellcolor{hcolor!20}H & \checkmark & Iterative & Param.& Time-to-event \\
FL-TTE~\cite{li2025federated}         & NPJ-DM'25 & \cellcolor{scorec!30}ATE     & IPW/Cox    & \cellcolor{hcolor!20}H & \checkmark & Iterative & Param.& Time-to-event \\
xFBCI~\cite{xiao2025bayesian}             & JIM'25    & \cellcolor{scorec!30}ATE      & Bayes./Match. & \cellcolor{hcolor!20}H & \checkmark & Iterative & SS & Continuous \\
Fed-IPW~\cite{khellaf2026federated}       & ICML'26  & \cellcolor{scorec!30}ATE      & IPW        & \cellcolor{hcolor!20}H & \checkmark & Iterative & SS    & Binary/Cont. \\
Fed-AIPW~\cite{khellaf2026federated}      & ICML'26  & \cellcolor{scorec!30}ATE      & AIPW       & \cellcolor{hcolor!20}H & \checkmark & Iterative & SS    & Binary/Cont. \\
FedIV~\cite{tyagi2025federated}         & arXiv'25  & \cellcolor{scorec!30}ATE      & IV/GMM     & \cellcolor{hcolor!20}H & \checkmark & Iterative & Param.& Continuous \\
Targeted DF~\cite{liu2025targeted}        & arXiv'25  & \cellcolor{scorec!30}ATE      & DR/Surv.   & \cellcolor{hcolor!20}H & \checkmark & Iterative & SS    & Time-to-event \\
ECO-ATE~\cite{li2026efficient}            & Biometrics'26  & \cellcolor{scorec!30}ATE      & DR         & \cellcolor{hcolor!20}H & \checkmark & 2 rounds  & SS    & Continuous \\
DC-DML~\cite{kawamata2026estimation}          & ESWA'26   & \cellcolor{lightgray}CATE     & DML        & \cellcolor{hcolor!20}H & --         & 1-shot    & Repr. & Continuous \\
PW-FedAvg~\cite{almodovar2023federated}   & NeurIPS-W'23 & \cellcolor{constraintc!20}ITE   & DL/VAE     & \cellcolor{hcolor!20}H & \checkmark & Iterative & Param.& Continuous \\
FedECE~\cite{zhao2025fedece}   & TAI'25 & \cellcolor{scorec!30}ATE   & CB/IDA     & \cellcolor{hcolor!20}H & \checkmark & Iterative & SS   & Continuous  \\
FedSIM~\cite{li2025sampling}   & arXiv'25 & \cellcolor{scorec!30}ATE   & M-est./MCMC     & \cellcolor{hcolor!20}H & \checkmark & 1-shot & SS   & Continuous  \\
FedCI-MSAE~\cite{khellaf2025federated_m}   & AISTATS'25 & \cellcolor{scorec!30}ATE   & G-formula/Meta     & \cellcolor{hcolor!20}H & \checkmark & 1-shot/Iter. & SS   & Continuous  \\
MVAgg~\cite{koga2024differentially}   & SaTML'24 & \cellcolor{scorec!30}ATE   & Match./DP     & \cellcolor{hcolor!20}H & \checkmark & 1-shot & DP   & Binary  \\
PFWS~\cite{cao2025heterogeneity}   & arXiv'25 & \cellcolor{constraintc!35}ATE/HTE   & DR/EIF     & \cellcolor{hcolor!20}H & \checkmark & 1-shot & SS   & Binary/Cont.  \\
PW FedAvg~\cite{almodovar2024propensity}   & CBM'24 & \cellcolor{constraintc!20}ITE   & DL/VAE     & \cellcolor{hcolor!20}H & \checkmark & Iterative & Param.   & Continuous  \\
TFLF~\cite{zhao2025new}   & arXiv'25 & \cellcolor{constraintc!35}ATE/HTE   & DR/EIF     & \cellcolor{hcolor!20}H & \checkmark & 1-shot & SS   & Binary/Cont.  \\
\bottomrule
\end{tabular}
\end{sidewaystable}
}

\subsection{ATE Estimation}\label{sec:fci_ate}

\subsubsection{IPW-Based Methods}

Inverse probability weighting (IPW) is a cornerstone of causal inference, and several federated methods extend it to distributed settings.

\textbf{FedMLE and FedIPW-MLE}~\cite{xiong2023federated} develop federated methods for heterogeneous observational data.
Locally computed propensity-score-based summary statistics are aggregated across sites to obtain point and variance estimators.
The authors identify that aggregation schemes must account for heterogeneity in both treatment assignments and outcomes across sites, and prove consistency and asymptotic normality.
The federated estimators are communication-efficient, requiring only one-way, one-time sharing of summary statistics.

\textbf{COLA}~\cite{hu2024collaborative} introduces a decentralized framework using sequential updating machinery for cross-site communication.
Unlike standard FL with a central coordinator, COLA operates peer-to-peer: the first site initializes, and subsequent sites refine the estimate.
Four algorithms (1r-cola through 4r-cola) trade off communication cost against statistical efficiency.
The authors prove that COLA achieves the same convergence rate as centralized analysis with the full cumulative sample size, a property not shared by standard meta-analysis.

\textbf{Fed-IPW and Fed-AIPW}~\cite{khellaf2026federated} propose a novel propensity score federation strategy using \emph{membership weights} (probabilities of site membership conditional on covariates) to construct a global propensity score as a mixture of local scores.
This allows sites to use different estimation methods locally and is particularly advantageous when overlap between treatment groups is poor or absent within individual sites.

\textbf{FedECA}~\cite{ogier2025fedeca} develops a federated external control arm method for \emph{time-to-event} outcomes using federated IPTW and Cox models.  Applied to metastatic pancreatic cancer data across three countries, it replicates pooled IPTW results up to machine precision while maintaining patient privacy.

\textbf{Limitations.}
Inverse probability weighting is only as good as the propensity-score model it relies on: misspecification of that model biases the estimate, and poor covariate overlap (near-violations of positivity) produces extreme weights that inflate variance.
These classical weaknesses persist in the federated setting and interact with it: one-shot schemes such as FedMLE and FedIPW-MLE~\cite{xiong2023federated} require each site to hold enough samples for its local propensity estimate to be reliable, and aggregating propensity models across sites with heterogeneous treatment-assignment mechanisms is delicate.
Strategies that explicitly model site membership, such as Fed-IPW and Fed-AIPW~\cite{khellaf2026federated}, alleviate poor within-site overlap but introduce an additional membership model that must itself be estimated and can be misspecified.

\subsubsection{Doubly Robust and Augmented Methods}

This family of methods brings augmented and doubly robust estimators, most notably AIPW, into the federated paradigm, inheriting their celebrated resilience to partial model misspecification while tackling the additional complexities that distributed healthcare networks introduce: covariate shift across sites, heterogeneous data distributions, communication bottlenecks, and stringent privacy requirements.

\textbf{FedAIPW}~\cite{xiong2023federated} extends the AIPW estimator to federated settings, enjoying double robustness.

\textbf{CLB-AIPW}~\cite{guo2024collaborative} proposes a \emph{collaborative} inverse propensity score weighting estimator that directly takes weighted means of nuisance models across sites rather than performing meta-analysis post-hoc.
This is the first method that allows collaboration across \emph{disjoint domains} without additional assumptions, and remains stable as heterogeneity increases.

\textbf{FACE}~\cite{han2025federated} develops a federated adaptive causal estimation framework with density ratio weighting to account for covariate shifts.
An adaptive penalized regression procedure guards against negative transfer, achieving both consistency and optimal efficiency.
Applied to comparing Pfizer and Moderna COVID-19 vaccines across five VA regional sites, it reduced standard errors by 26--67\% compared to traditional methods.

\textbf{MR\_L1}~\cite{han2023multiply} offers a multiply-robust, privacy-preserving estimator that accommodates \emph{covariate mismatch} (different covariates across sites) using transfer learning for ensemble weights.

\textbf{RIFL}~\cite{guo2025robust} provides robust inference for the \emph{prevailing model} (matching the majority of sites) using a novel resampling method that accounts for site selection uncertainty.
Applied to mortality risk prediction across 274 hospitals in four countries using COVID-19 EHR data.

\textbf{ECO-ATE}~\cite{li2026efficient} achieves the semiparametric efficiency bound under data-sharing constraints with only two rounds of communication, allowing distributional shifts across sites.

\textbf{FedCI-MSAE}~\cite{khellaf2025federated_m} systematically compares three classes of federated ATE estimators derived from the Plug-in G-Formula for randomized controlled trials: (i)~meta-analysis estimators that aggregate independently computed per-study ATE estimates, (ii)~one-shot federated estimators that aggregate outcome model parameters in a single round, and (iii)~gradient-based federated estimators that learn the outcome model on joint data via federated gradient descent.
The authors derive asymptotic variances under linear outcome models and show that meta-analysis estimators can achieve statistical efficiency comparable to pooled analysis when local samples are sufficient, while gradient-based estimators are necessary when local datasets are small.
One-shot estimators offer an intermediate trade-off, recovering pooled ATE estimates under distributional shifts but suffering from increased variance under study-effects.

\textbf{PFWS}~\cite{cao2025heterogeneity} develops semiparametrically efficient estimators for a broad class of causal measures (including risk ratios) across multiple federated sources under two transportability assumptions with different efficiency-robustness trade-offs.
The estimators support flexible machine learning for nuisance functions while maintaining parametric convergence rates.
To handle scenarios where some source sites violate transportability, a \emph{Post-Federated Weighting Selection} (PFWS) framework adaptively identifies compatible sites via a two-step procedure, asymptotically achieving the semiparametric efficiency bound while mitigating negative transfer.

\textbf{TFLF}~\cite{zhao2025new} introduces a targeted-federated learning framework for estimating heterogeneous treatment effects (HTEs) for a prespecified target population.
It defines HTEs via a projection-based estimand applicable to both continuous and binary outcomes, and develops a doubly robust estimator that integrates information from multiple data sources while accounting for covariate distribution shifts through density ratio modeling.
A communication-efficient bootstrap-based selection procedure detects and excludes non-transportable data sources, requiring only a single round of information exchange.
Applied to nationwide Medicare-linked data for comparing hip fracture surgical treatments.

\textbf{Limitations.}
Double robustness guarantees consistency only when at least one of the two nuisance models (outcome regression or propensity score) is correctly specified; if both are misspecified the estimator is biased, and through its weighting component it remains sensitive to poor overlap.
These methods are also more involved to deploy, since both nuisance models must be estimated at every site.
Moreover, the transfer-learning and site-selection mechanisms that several of them use to guard against negative transfer (\eg, FACE~\cite{han2025federated}, RIFL~\cite{guo2025robust}, PFWS~\cite{cao2025heterogeneity}, and TFLF~\cite{zhao2025new}) rest on transportability assumptions that may not hold in practice, and the data-driven selection step introduces additional uncertainty that is not always reflected in the reported confidence intervals.

\subsubsection{Matching-Based Methods}

Rather than modeling outcomes or reweighting observations, these methods extend classical matching techniques to federated environments, pairing treated and control units across distributed sites on the basis of estimated propensity scores while incorporating privacy mechanisms such as differential privacy and leveraging federated model training for score estimation.

\textbf{FPSM}~\cite{liu2025federated_p} extends propensity score matching to FL using a \emph{Federated Random Forest} (FRF) for propensity estimation, federated nearest-neighbor matching, and federated ATE estimation.
It further combines with DID analysis (FPSM-DID) for dynamic panel data.

\textbf{xFBCI}~\cite{xiao2025bayesian} integrates Bayesian federated learning with propensity score matching for manufacturing applications (EHD printing), using Expectation Propagation and Stochastic Gradient Langevin Dynamics for scalable posterior estimation.

\textbf{MVAgg}~\cite{koga2024differentially} addresses multi-site ATE estimation with per-site differential privacy guarantees.
For observational studies, it proposes \emph{SmoothDPMatching}, a smooth-sensitivity-based DP matching algorithm that adds significantly less noise than naive global sensitivity baselines.
Each site reports both its private ATE estimate and a private variance estimate as a quality measure.
On the server side, the \emph{minimum-variance aggregation} (MVAgg) algorithm selects an optimal subset of sites to aggregate, minimizing the overall variance of the final ATE estimate.
This three-component design (per-site DP estimation, quality reporting, and optimal aggregation) effectively handles site heterogeneity in sample sizes and privacy budgets.

\textbf{Limitations.}
Matching estimators discard units that cannot be paired, reducing the effective sample size and statistical efficiency, and their quality degrades sharply when covariate overlap between treated and control groups is limited.
The federated setting compounds this, because units at different sites cannot be matched directly without exposing individual records, forcing reliance on propensity-score summaries or reduced-dimension representations that may match only coarsely.
Privacy mechanisms add a further tension: the differential-privacy noise injected by MVAgg~\cite{koga2024differentially}, although minimized through smooth sensitivity, still trades estimation accuracy for protection.

\subsubsection{Data Collaboration Methods}
These methods permit sites to share carefully constructed intermediate representations, summary statistics, or synthetic data, balancing the statistical benefits of pooled information against the privacy constraints that preclude direct data exchange.

\textbf{DC-QE}~\cite{kawamata2024collaborative} proposes a one-shot method using dimensionality-reduced intermediate representations shared across parties. It uniquely addresses the lack of both subjects and covariates across parties, supporting both IPW and matching estimation.

\textbf{DC-DML}~\cite{kawamata2026estimation} extends this to semi-parametric CATE estimation via double machine learning on privacy-preserving fusion data, enabling collaborative estimation across time points and parties.

\textbf{Limitations.}
By sharing dimensionality-reduced representations rather than raw data, these methods reduce communication and exposure, but the reduction is lossy: information discarded during compression cannot be recovered, and the validity of the downstream estimate hinges on the assumption that the shared representations preserve the confounding structure relevant to the target effect.
The privacy they afford is also informal, since reduced-dimension features can still leak information under reconstruction or membership-inference attacks unless paired with formal guarantees.

\subsection{ITE and CATE Estimation}\label{sec:fci_ite}

\subsubsection{Bayesian, Gaussian Process, and Kernel Methods}
This family adopts a probabilistic or nonparametric view of treatment effect estimation.
Bayesian and Gaussian-process approaches place prior distributions over potential outcomes or model parameters and update them through site-level likelihoods, achieving principled uncertainty quantification and natural aggregation of heterogeneous beliefs without direct data sharing; kernel and random-feature approaches instead ground estimation in reproducing kernel Hilbert spaces, or their finite-dimensional random-feature approximations, offering flexible nonparametric modeling of treatment effect functions while keeping communication overhead tractable.
What unites them is that per-site contributions enter the global objective additively, making them naturally amenable to federated decomposition.

\textbf{FedCI}~\cite{vo2022bayesian} proposes a Bayesian framework using Gaussian processes to model potential outcomes, with a variational approximation whose evidence lower bound decomposes additively across sources, enabling federated gradient averaging and posterior uncertainty quantification.

\textbf{CausalFI}~\cite{vo2025federated} extends this to handle dissimilar distributions across sources and missing data under MAR/MCAR assumptions by recovering conditional distributions of missing confounders.

\textbf{CausalRFF}~\cite{vo2022adaptive} leverages Random Fourier Features (RFF) to approximate kernel functions, naturally inducing decomposition of the loss function into per-source components.
Adaptive kernel functions handle dissimilar distributions, and the authors provide minimax lower bounds.

\textbf{Limitations.}
These estimators provide principled uncertainty quantification but inherit the scalability burden of their building blocks: exact Gaussian-process inference scales cubically with sample size, so FedCI~\cite{vo2022bayesian} and CausalFI~\cite{vo2025federated} rely on variational approximations whose error is hard to quantify, while CausalRFF~\cite{vo2022adaptive} attains tractability only by approximating the kernel with random features.
Their conclusions are further sensitive to the choice of prior and kernel, and the additive cross-site decomposition that enables federation rests on a conditional-independence assumption across sites that can be violated when sites share unobserved common causes.

\subsubsection{Deep Generative and Transformer Methods}

Leveraging the expressive capacity of generative models and attention-based architectures, these approaches learn rich confounder and treatment effect representations across federated sites through privacy-preserving training and cross-site alignment strategies.

\textbf{FedTEDVAE}~\cite{almodovar2023federated} applies FedAvg to TEDVAE (Treatment Effect with Disentangled VAE) for federated ITE estimation, demonstrating the feasibility of combining federated learning with advanced deep generative causal models.
\textbf{PW FedAvg}~\cite{almodovar2024propensity} extends this by addressing the \emph{propensity score shift} problem -- the phenomenon where treatment assignment criteria differ across hospitals (e.g., due to varying drug availability).
It adapts the FedAvg aggregation process by weighting each client's contribution proportionally to the number of samples trained on, mitigating biases from imbalanced treatment assignment distributions.
Experiments on semi-synthetic causal inference benchmarks show that PW FedAvg bridges the gap between centralized and isolated training, outperforming vanilla FedAvg and other distributed causal inference baselines (linear models, Gaussian Processes, Random Fourier Features) as treatment imbalance increases across nodes.

\textbf{FedTransTEE}~\cite{makhija2024federated} is the first transformer-based federated ITE estimation framework, handling heterogeneous covariates, treatments, and outcome spaces across sites through cross-attention mechanisms and intervention embeddings that enable zero-shot estimation for unseen treatments.

\textbf{Fed-IPTW}~\cite{yin2026federated} addresses treatment assignment strategy variation across hospitals via two-step FedAvg: first learning patient-specific propensity weights for local decorrelation, then learning unbiased factual predictions with hospital-specific weights for global decorrelation.

\textbf{Limitations.}
Deep generative and attention-based estimators are data-hungry, so the limited per-site sample sizes typical of federated deployments can undermine the very representational flexibility that motivates them.
Unlike summary-statistic methods, they generally exchange model parameters or gradients across rounds, which raises communication cost and exposes a larger attack surface to inference attacks unless paired with explicit privacy mechanisms.
They also offer weaker theoretical guarantees, with formal identifiability and consistency results largely absent, and their black-box nature makes the effect estimates harder to interpret and audit than those of semiparametric estimators.

\subsubsection{Semiparametric and Specialized Estimators}

Beyond the probabilistic and deep-learning families above, several methods bring \emph{specialized semiparametric estimators} into the federated setting, each addressing a distinct source of difficulty.
What they share is a reliance on estimating equations or orthogonalized scores whose per-site components can be aggregated without pooling raw data; they differ in the estimand and nuisance structure they target: high-dimensional confounding via double machine learning, time-to-event outcomes via survival models, unmeasured confounding via instrumental variables, and non-smooth objectives via sampling-based inference.

\emph{Double machine learning.}
The Neyman orthogonality principle anchors this line of work: cross-fitted nuisance estimators partial out high-dimensional confounders, isolating the target causal parameter in a way that is both robust to nuisance estimation error and naturally suited to federated decomposition.
\textbf{FDML}~\cite{kang2025federated} extends double machine learning to FL for high-dimensional semiparametric models, using surrogate Neyman-orthogonal scores and density ratio tilting, and is applied to multi-phase ADNI data for Alzheimer's disease research.

\emph{Survival analysis.}
Targeting time-to-event outcomes, these methods adapt classical survival models such as the Cox proportional hazards framework to federated settings, carefully handling censoring, site-level heterogeneity, and the privacy-sensitive aggregation of risk sets.
\textbf{FL-TTE}~\cite{li2025federated} proposes federated target trial emulation for time-to-event outcomes, combining federated propensity scoring with federated Cox proportional hazards models, validated on sepsis (192 hospitals) and Alzheimer's trials (5 health systems).
\textbf{Targeted Data Fusion}~\cite{liu2025targeted} develops both a semiparametric efficient estimator (when data can be shared) and a federated learning framework (under privacy constraints) for survival analysis, dynamically reweighting source contributions.

\emph{Instrumental variables.}
When unmeasured confounding renders standard estimators inconsistent, these methods restore identifiability by exploiting exogenous instruments, extending two-stage procedures to federated configurations where instruments and outcomes may reside at non-overlapping sites.
\textbf{FedIV}~\cite{tyagi2025federated} introduces federated instrumental variable analysis via federated GMM, formulated as a federated zero-sum game, and establishes that equilibrium solutions consistently estimate local moment conditions of every client, even under heterogeneity.

\emph{Sampling-based inference.}
Intractable posteriors over causal quantities are approximated through Monte Carlo or variational sampling schemes distributed across sites, with privacy preserved via message-passing or likelihood-sharing protocols that avoid direct exposure of local data.
\textbf{FedSIM}~\cite{li2025sampling} develops a sampling-based federated learning framework for statistical inference on M-estimators with non-smooth objective functions, which arise in applications such as quantile regression and AUC maximization.
The method uses Markov Chain Monte Carlo (MCMC) sampling with a second-stage perturbation scheme to efficiently estimate both parameters and their variances, circumventing the need for nonparametric estimation of nuisance quantities.
For multi-site settings, an adaptive strategy borrows information from potentially heterogeneous source sites via a dissimilarity-based selection mechanism and a lasso-regularized weighting scheme.
The resulting estimator possesses an oracle property: it achieves optimal asymptotic efficiency by leveraging eligible sites while guarding against negative transfer, requiring only one round of communication without sharing individual-level data.

\textbf{Limitations.}
Each specialized estimator inherits the assumptions of the classical machinery it federates.
Double machine learning (FDML~\cite{kang2025federated}) needs high-quality cross-fitted nuisance estimates and hence adequate local samples; survival methods (FL-TTE~\cite{li2025federated}, Targeted Data Fusion~\cite{liu2025targeted}) lean on modeling choices such as proportional hazards and on the privacy-sensitive aggregation of risk sets; and instrumental-variable analysis (FedIV~\cite{tyagi2025federated}) restores identifiability only under the strong, largely untestable assumptions of instrument validity, with weak instruments inflating variance.
Sampling-based inference (FedSIM~\cite{li2025sampling}) avoids nuisance estimation but at the price of the computational and convergence-diagnostic burden of MCMC.
Across this group, theoretical guarantees are typically established under specific heterogeneity and overlap conditions that may not hold at every site.


\section{Connections Between FCD and FCI}\label{sec:connection}

\begin{figure}[t!]
	\centering
	\includegraphics[width=1.0\linewidth]{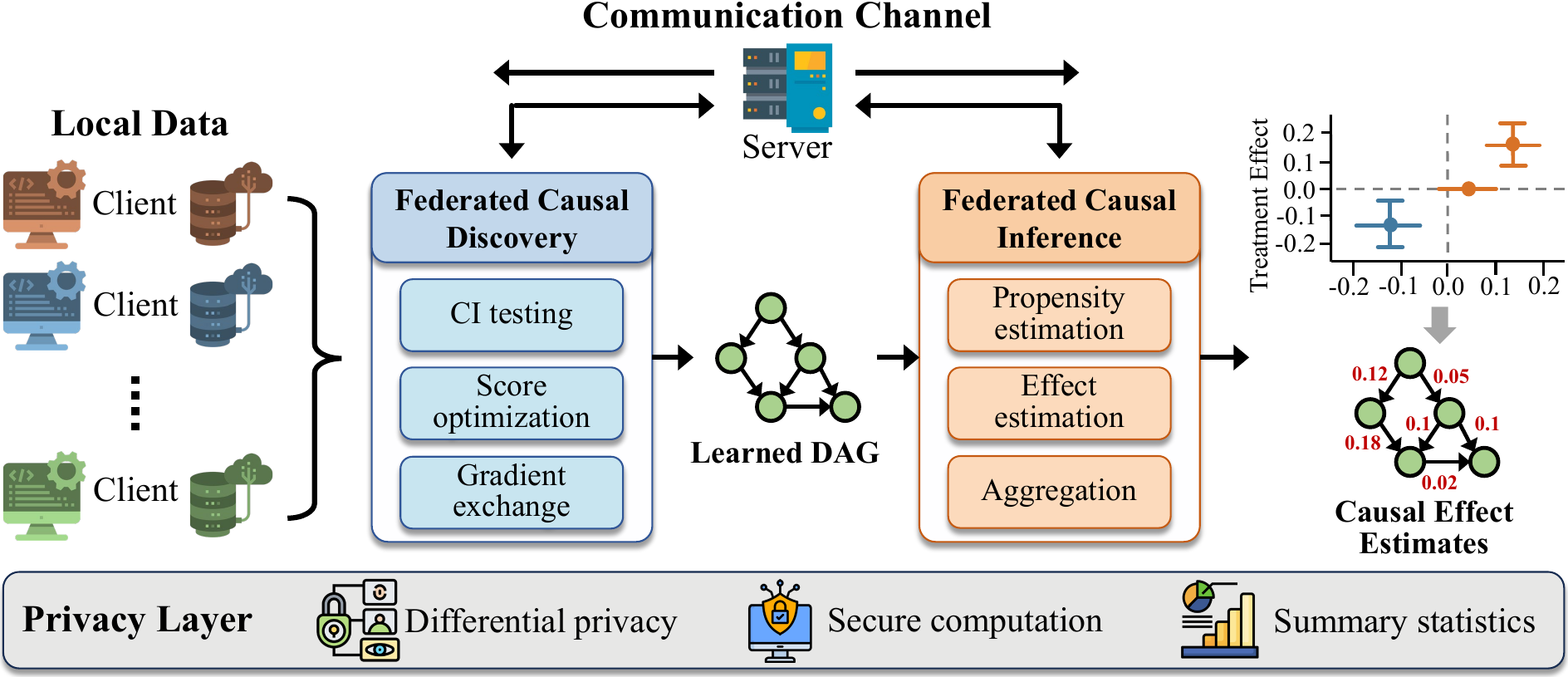}
	\caption{The pipeline connecting federated causal discovery and federated causal inference.}
	\label{fig:pipeline}
\end{figure}

Federated causal discovery and federated causal inference, while often studied separately, are deeply interconnected, as illustrated in the unified pipeline shown in Figure~\ref{fig:pipeline}.
The two tasks answer complementary questions over the same distributed data: FCD asks \emph{which} variables cause \emph{which}, recovering the structure $\mathcal{G}$, whereas FCI asks \emph{how much} a designated treatment changes a designated outcome, quantifying an effect such as the ATE $\tau$.
Crucially, the validity of any FCI estimate hinges on structural assumptions, that is, which variables must be adjusted for, which may serve as instruments, and which lie on mediating pathways, that are precisely the objects FCD is designed to recover.
This makes FCD and FCI not merely related but sequentially dependent: structure is the input to identification, and identification is the prerequisite for unbiased estimation.

While prior work~\cite{deng2026survey} discusses FL for causality in general, to the best of our knowledge, this is the first work to explicitly formalize and analyze the interplay between federated causal discovery and federated causal inference as complementary stages of a unified federated causal reasoning pipeline, with formal problem definitions linking the two.
In the remainder of this section, we make this connection precise, surveying how the two stages compose, where their machinery is shared, and what new difficulties arise specifically at their interface.

\subsection{A Unified Problem Formulation}\label{sec:connection_formulation}

Consider $K$ clients, where client $k$ holds a private dataset $\mathcal{D}_k$ over a common (or partially overlapping) variable set $\mathbf{V}$ that includes a designated treatment $A$ and outcome $Y$.
We formalize federated causal reasoning as the composition of two privacy-preserving stages.

\textbf{Stage 1 (Federated structure learning).}
The clients collaboratively recover a global causal structure by exchanging only privacy-preserving statistics (CI-test summaries, scores, or gradients) rather than raw records,
\begin{equation}
    \hat{\mathcal{G}} = \mathcal{A}_{\mathrm{FCD}}\big(\{\mathcal{D}_k\}_{k=1}^{K}\big),
    \label{eq:stage1}
\end{equation}
where $\hat{\mathcal{G}}$ is a DAG or, more commonly, a CPDAG representing its Markov equivalence class (MEC).

\textbf{Stage 2 (Federated effect estimation).}
Given the learned structure $\hat{\mathcal{G}}$ and the target pair $(A,Y)$, the clients first \emph{identify} a valid adjustment set $\hat{\mathbf{Z}} \subseteq \mathbf{V}\setminus\{A,Y\}$ from $\hat{\mathcal{G}}$, for example via the back-door criterion~\cite{pearl2009causality}, and then collaboratively estimate the effect,
\begin{equation}
    \hat{\tau} = \mathcal{A}_{\mathrm{FCI}}\big(\{\mathcal{D}_k\}_{k=1}^{K};\, A, Y, \hat{\mathbf{Z}}\big).
    \label{eq:stage2}
\end{equation}
The composition $\hat{\tau} = \big(\mathcal{A}_{\mathrm{FCI}} \circ \mathcal{A}_{\mathrm{FCD}}\big)(\{\mathcal{D}_k\})$ defines the \emph{federated causal reasoning pipeline}.
This formulation exposes two facts that motivate the rest of the section: the unconfoundedness assumption underlying Eq.~\eqref{eq:adjustment} is \emph{not} given a priori but must be supplied by Stage 1, and any error or ambiguity in $\hat{\mathcal{G}}$ propagates into $\hat{\tau}$.

\subsection{FCD as an Upstream Step for FCI}\label{sec:connection_upstream}

The most direct use of the pipeline is to let FCD furnish the structural knowledge that FCI requires.
A learned DAG identifies valid adjustment sets via the back-door criterion, exposes candidate instrumental variables, and reveals mediating pathways, each of which improves the reliability and interpretability of the downstream effect estimate.
This data-driven identification is especially valuable in the federated setting, where no single analyst has global visibility into the joint distribution and where hand-specifying a confounder set across heterogeneous sites is error-prone.

\textbf{FedECE}~\cite{zhao2025fedece} is the most explicit realization of this pipeline: it proposes a unified framework comprising a federated causal structure learning (FedCSL) module and a federated causal effect (FedCE) module.
The FedCSL module uses a layer-wise cooperative optimization strategy to learn a global skeleton and a distributed optimal consensus strategy for V-structure identification, all without sharing raw data.
The FedCE module then computes causal effects based on the learned CPDAG using a progressively integrated multiset strategy that handles the inherent ambiguity of undirected edges.
Three instantiations are proposed and validated on synthetic, benchmark, and real datasets, including the IHDP dataset: FedECE-B (basic), FedECE-L (computationally efficient), and FedECE-O (accuracy-optimized).
Beyond FedECE, xFBCI~\cite{xiao2025bayesian} also explicitly uses a Bayesian network structure to derive propensity scores for matching, while several other FCI methods implicitly assume a known or partially known causal graph and therefore depend, often silently, on an upstream discovery step that an FCD method could supply.

\subsection{Sequential versus Joint Integration}\label{sec:connection_integration}

Two paradigms can realize the composition in Eqs.~\eqref{eq:stage1}--\eqref{eq:stage2}.
The \emph{sequential} (two-stage) paradigm runs FCD to completion, fixes $\hat{\mathcal{G}}$, and then runs FCI conditioned on it; FedECE exemplifies this modular design.
Its advantages are practical: the two stages can reuse existing federated primitives, be audited independently, and respect separate privacy budgets.
Its drawback is that errors in $\hat{\mathcal{G}}$ are treated as ground truth downstream, so structural mistakes translate directly into biased effects.
The \emph{joint} (end-to-end) paradigm instead optimizes a single objective that couples structure learning and effect estimation, for instance by sharing nuisance representations between the two stages or by regularizing the discovery loss with downstream estimation accuracy.
Joint optimization can, in principle, achieve better statistical efficiency by allowing the target estimand to inform which structural ambiguities are worth resolving, but it complicates privacy accounting and federated optimization and remains largely unexplored.
Characterizing when the modular and joint paradigms differ, and by how much, is an open theoretical question at the heart of this connection.

\subsection{Propagating Structural Uncertainty}\label{sec:connection_uncertainty}

A defining feature of the pipeline is that Stage 1 typically identifies the structure only up to a Markov equivalence class: from observational data alone, $\hat{\mathcal{G}}$ is a CPDAG with undirected edges whose orientation is not determined.
Consequently, a single dataset is consistent with multiple adjustment sets and hence multiple possible effect values.
Classical centralized treatments such as IDA~\cite{maathuis2009estimating} address this by enumerating the DAGs in the MEC and returning the \emph{set} of possible causal effects rather than a single number; FedECE's progressively integrated multiset strategy is the federated analogue, propagating CPDAG ambiguity through to a multiset of effect estimates.
Naively reporting a point estimate from one arbitrarily chosen member of the equivalence class can yield overconfident conclusions.
Two challenges are therefore specific to the federated interface: (i)~propagating the equivalence-class ambiguity of $\hat{\mathcal{G}}$ into a calibrated uncertainty band on $\hat{\tau}$, and (ii)~accounting for the additional finite-sample uncertainty in $\hat{\mathcal{G}}$ that arises because each CI decision or score is itself an aggregate over noisy, heterogeneous local statistics.
Federated sensitivity analysis~\cite{ding2016sensitivity}, quantifying how robust $\hat{\tau}$ is to residual structural uncertainty and to possible violations of causal sufficiency, is a natural but underdeveloped tool here.

\subsection{From Inference Back to Discovery}\label{sec:connection_reverse}

The dependency is not one-directional.
Federated discovery methods that can reliably orient edges, especially those that exploit interventional data such as FedCDI~\cite{abyaneh2022federated} or unknown client-level interventions such as I-PERI~\cite{baldo2026regret}, shrink the equivalence class and thereby narrow the space of possible effect estimands, making downstream inference both more targeted and more efficient.
Conversely, the inference stage can feed back into discovery: a designated treatment--outcome pair imposes constraints (e.g., a known causal direction $A \to Y$, or background knowledge that certain variables are pre-treatment) that can orient otherwise-undirected edges in $\hat{\mathcal{G}}$, and effect-based criteria can prioritize which structural ambiguities are worth the communication cost of resolving.
Treating the pipeline as a loop rather than a chain, in which preliminary effect estimates guide subsequent rounds of targeted structure refinement, is a promising direction that no existing federated method fully realizes.

\subsection{Shared Machinery and Open Challenges}\label{sec:connection_shared}

Beyond their logical dependency, FCD and FCI confront the same federated difficulties and can reuse one another's solutions.
Both must handle data heterogeneity across sites, preserve privacy while retaining statistical utility, control communication cost, and cope with non-identical variable sets.
The technical primitives also transfer: membership-weight-based propensity-score aggregation from Fed-IPW~\cite{khellaf2026federated} suggests how to aggregate CI-test statistics across heterogeneous sites in federated discovery, while the secure federated CI tests developed for FCD~\cite{huang2023towards,li2024federated} can supply the conditional-independence checks that identification in FCI relies upon.
A distinctive challenge emerges only when the stages are composed: \emph{privacy budgets must be accounted for jointly}.
Under differential privacy, the sequential composition of a discovery mechanism and an estimation mechanism degrades the overall guarantee~\cite{dwork2014algorithmic}, so a naive concatenation of two separately calibrated methods may exceed the intended budget; designing end-to-end pipelines whose total privacy loss is controlled, while preserving both structural and estimation accuracy, is a central open problem.
Together, these observations show that the connection between FCD and FCI is not a convenience of presentation but a source of concrete research questions, spanning identification, uncertainty quantification, optimization, and privacy, that are invisible when the two tasks are studied in isolation.

\section{Applications}\label{sec:apps}

The integration of federated learning with causal reasoning has found impactful applications across diverse domains where data privacy, institutional boundaries, and heterogeneous data distributions pose fundamental challenges. In this section, we provide a detailed discussion of representative application domains, highlighting how FCD and FCI methods address domain-specific requirements.

\subsection{Healthcare and Medicine}

Healthcare is the dominant application domain, driven by strict privacy regulations (e.g., HIPAA in the United States, GDPR in Europe) and the natural distribution of patient data across hospitals and health systems.

\textbf{Problem formulation.}
Consider $K$ hospitals, where hospital $k$ holds patient records $\mathcal{D}_k = \{(\mathbf{X}_i^{(k)}, A_i^{(k)}, Y_i^{(k)})\}_{i=1}^{n_k}$, with covariates $\mathbf{X}$ (demographics, comorbidities, lab values), treatment $A$ (e.g., surgical procedure, drug assignment), and outcome $Y$ (e.g., survival time, recovery status).
The goal is to estimate the causal effect $\tau = \mathbb{E}[Y(1) - Y(0)]$ or the heterogeneous treatment effect $\tau(\mathbf{x}) = \mathbb{E}[Y(1) - Y(0) \mid \mathbf{X} = \mathbf{x}]$ without sharing individual patient records.
A key challenge is that treatment assignment policies may differ across hospitals, inducing heterogeneous propensity scores $e_k(\mathbf{x}) = P(A=1 \mid \mathbf{X}=\mathbf{x}, S=k)$ where $S$ denotes the site indicator.

\textbf{Federated causal discovery applications.}
Federated causal discovery enables the recovery of disease mechanisms and gene regulatory networks across medical institutions.
NOTEARS-PFL~\cite{liu2025federated} applies personalized federated BN learning to multi-site neuroimaging data for major depressive disorder, recovering both shared and site-specific functional connectivity patterns.
Fed-HC-aIPW~\cite{zanga2025federated} addresses causal structure learning in endometrial cancer studies across medical centers, handling heterogeneous missing data mechanisms.

\textbf{Federated causal inference applications.}
RIFL~\cite{guo2025robust} applies robust federated inference to COVID-19 mortality risk prediction across 274 hospitals in four countries, using electronic health record (EHR) data with substantial cross-site heterogeneity.
FACE~\cite{han2025federated} compares Pfizer and Moderna COVID-19 vaccine effectiveness across five Veterans Affairs regional sites, achieving 26--67\% reductions in standard errors compared to traditional meta-analytic approaches.
FDML~\cite{kang2025federated} and FL-TTE~\cite{li2025federated} apply federated causal inference to Alzheimer's disease treatment effect estimation using multi-phase ADNI data and time-to-event outcomes, respectively.
FedECA~\cite{ogier2025fedeca} develops federated external control arms for metastatic pancreatic cancer treatment comparison across three countries.
Additional applications include sepsis management~\cite{li2025federated}, post-transplantation diabetes prevention~\cite{hu2024collaborative}, hip fracture surgery outcomes~\cite{liu2025targeted}, and acute myocardial infarction treatment~\cite{han2023multiply}.
Rocchi~\etal~provide a comprehensive perspective on the trends, opportunities, and challenges of FCD, specifically in medicine~\cite{rocchi2025federated}.

\subsection{Manufacturing and Engineering}

In smart manufacturing and distributed engineering systems, production data is often siloed across factories, production lines, or equipment vendors due to proprietary concerns.

\textbf{Problem formulation.}
Let $\mathbf{X}^{(k)} \in \mathbb{R}^{n_k \times d_k}$ denote the process parameter measurements at facility $k$, and let $Y^{(k)}$ denote the quality outcome.
The federated causal discovery task aims to learn the causal DAG $\mathcal{G}^*$ over process parameters to identify root causes of quality defects, while the federated causal inference task estimates the interventional distribution $P(Y \mid do(\mathbf{X}_j = x))$ to optimize process settings.

xFBCI~\cite{xiao2025bayesian} applies federated Bayesian causal inference to electrohydrodynamic (EHD) printing optimization and smart manufacturing maintenance, using Expectation Propagation and Stochastic Gradient Langevin Dynamics for scalable posterior estimation across distributed production data.
FedGCL~\cite{mohanty2025federated} and FedGC~\cite{mohanty2026uncertainty} target distributed infrastructure such as smart grids and cyber-physical systems, where federated Granger causality learning enables anomaly root-cause analysis without centralizing sensitive operational data. The uncertainty quantification framework in FedGC is particularly valuable for safety-critical engineering applications~\cite{pearl2009causality}.

\subsection{Social Media and Digital Platforms}

Social media platforms generate vast amounts of distributed, privacy-sensitive user interaction data across content moderation systems, recommendation engines, and advertising platforms.

CCaT~\cite{tregubov2024tool} demonstrates collaborative causal discovery for social media content moderation, where multiple platform divisions jointly learn how content features causally affect user engagement and policy violations, sharing only summary causal relations rather than user-level data.

\textbf{Recommendation systems.}
Causal reasoning in federated recommendation systems addresses the confounding between user preferences and exposure mechanisms~\cite{wang2021deconfounded,zhang2021causal}.
Formally, let $\mathbf{U}$ denote user features, $\mathbf{I}$ denote item features, and $Y$ denote the outcome (e.g., click, purchase).
The exposure mechanism $P(\text{exposed} \mid \mathbf{U}, \mathbf{I})$ confounds the observed association between user-item features and outcomes.
Federated causal methods can deconfound this relationship across distributed platforms by estimating the causal effect $\mathbb{E}[Y \mid do(\mathbf{I}), \mathbf{U}]$ without pooling user data, leveraging techniques from federated propensity estimation~\cite{xiong2023federated} and causal representation learning~\cite{scholkopf2021toward}.

\subsection{Economics and Policy Evaluation}

Causal effect estimation is central to economic research and policy evaluation, where data is often distributed across government agencies, financial institutions, or regional statistical offices~\cite{imbens2015causal,athey2017state}.

\textbf{Problem formulation.}
Consider the evaluation of a policy intervention across $K$ administrative regions.
Region $k$ holds data $\mathcal{D}_k = \{(\mathbf{X}_i^{(k)}, A_i^{(k)}, Y_i^{(k)})\}_{i=1}^{n_k}$, where $A$ indicates policy exposure and $Y$ is the economic outcome (\eg, employment, income, firm productivity).
Under a difference-in-differences (DID) framework, the target estimand is $\tau_{\text{DID}} = \mathbb{E}[Y_{t=1}(1) - Y_{t=1}(0)] - \mathbb{E}[Y_{t=0}(1) - Y_{t=0}(0)]$, which requires parallel trends across pre- and post-intervention periods.

FPSM-DID~\cite{liu2025federated_p} addresses dynamic panel data in economic and policy evaluation settings by combining federated propensity score matching with DID analysis, enabling cross-site causal evaluation without data pooling.
FedIV~\cite{tyagi2025federated} introduces federated instrumental variable analysis for consumer economics applications, handling settings where unobserved confounding invalidates standard adjustment strategies but valid instruments are available at distributed sites.
DC-DML~\cite{kawamata2026estimation} enables semi-parametric CATE estimation across multiple time points and parties via double machine learning on privacy-preserving fusion data, offering a flexible framework for heterogeneous policy effect estimation~\cite{chernozhukov2018double}.

\subsection{Social Equity and Fairness}

Federated causal methods are increasingly relevant for auditing and promoting fairness across distributed decision-making systems (\eg, lending, hiring, criminal justice), where sensitive individual data cannot be centralized~\cite{mehrabi2021survey,kusner2017counterfactual}.

\textbf{Problem formulation.}
Let $S$ denote a sensitive attribute (\eg, race, gender), $\mathbf{X}$ denote legitimate covariates, and $Y$ denote the decision outcome.
Causal fairness notions~\cite{kilbertus2017avoiding,nabi2018fair} require estimating path-specific effects:
\begin{equation}
\text{NDE}_{S \to Y} = \mathbb{E}[Y(s, \mathbf{M}(s')) - Y(s', \mathbf{M}(s'))],
\label{eq:nde}
\end{equation}
where $\mathbf{M}$ denotes mediating variables and NDE is the natural direct effect.
In a federated setting, the challenge is to audit whether $\text{NDE}_{S \to Y} = 0$ (no direct discrimination) across institutions without revealing individual-level sensitive attributes.
Federated causal discovery can identify the pathways through which sensitive attributes affect outcomes~\cite{li2024federated,deng2026survey}, while federated causal inference methods can quantify these effects under privacy constraints.

\section{Challenges and Future Directions}\label{sec:future}

Despite significant progress, numerous open challenges remain at the intersection of federated learning and causal reasoning. In this section, we discuss these challenges in depth, highlighting formal problem formulations and promising research directions.

\subsection{Formal Privacy Guarantees}

While many FCD methods claim to be ``privacy-preserving'' by not sharing raw data, the exchange of summary statistics, model parameters, or even graph structures can leak information through inference attacks~\cite{zhu2019deep_gradient}.
Only a few methods provide formal privacy guarantees: PERI and I-PERI offer $\epsilon$-differential privacy (DP)~\cite{mian2023nothing,baldo2026regret}, and Fed-Sparse-BNSL provides $(\epsilon,\delta)$-DP~\cite{fehri2026differentially}.

\textbf{Open problems.}
A fundamental tension exists between privacy and causal identifiability. Formally, let $\mathcal{M}$ denote a privacy mechanism that perturbs shared statistics.
The privacy-utility trade-off can be expressed as:
\begin{equation}
    \min_{\mathcal{M}} \; \mathbb{E}\left[\Delta(\hat{\mathcal{G}}_{\mathcal{M}}, \mathcal{G}^*)\right] \quad \text{s.t.} \quad \mathcal{M} \text{ satisfies } (\epsilon, \delta)\text{-DP},
    \label{eq:privacy_tradeoff}
\end{equation}
where $\Delta(\cdot,\cdot)$ is a structural distance metric (e.g., structural Hamming distance) and $\mathcal{G}^*$ is the true causal graph.
Characterizing the minimax rate of this trade-off -- i.e., the fundamental limits of structure learning accuracy under privacy constraints~\cite{dwork2014algorithmic} -- remains an open problem.
Extending rigorous privacy analysis to constraint-based methods (where CI test statistics are shared) and continuous-optimization methods (where gradient information is exchanged), and developing tight composition theorems for multi-round causal protocols, are important directions.
Additionally, integrating secure multi-party computation~\cite{knott2021crypten} with differential privacy for stronger protection deserves further investigation.

\subsection{Communication Efficiency}

Most continuous-optimization-based FCD methods require exchanging $d \times d$ adjacency matrices (or gradients thereof) per iteration, leading to $O(d^2)$ communication cost per round, which scales poorly with dimensionality.
Fed-Sparse-BNSL's sparse-update approach~\cite{fehri2026differentially} is a promising exception, transmitting only $O(s)$ non-zero entries where $s \ll d^2$ is the sparsity level.

\textbf{Open problems.}
Developing communication-efficient protocols is critical for large-scale deployment.
Promising approaches include: (i) gradient compression and quantization techniques~\cite{bernstein2018signsgd} adapted to the structural constraints of DAG learning; (ii) sketching methods that preserve CI test statistics or score functions within bounded communication budgets; (iii) one-shot aggregation strategies inspired by communication-efficient federated optimization~\cite{zhang2022dense}; and (iv) event-triggered communication that exchanges updates only when local structural changes exceed a significance threshold.
For FCI, the communication cost is generally lower (many methods require only one-shot summary statistic exchange), but as methods grow more sophisticated (\eg, iterative doubly robust estimation), communication-accuracy trade-offs need formal analysis.

\subsection{Theoretical Foundations}

Convergence guarantees exist for only a few FCD methods: DARLS~\cite{ye2024federated} proves convergence at rate $O(\log(n)/\sqrt{m})$, Fed-Sparse-BNSL provides convergence under DP constraints, and FDBNL~\cite{chen2025federated} offers the first convergence analysis for continuous-optimization-based structure learning in FL.

\textbf{Open problems.}
Fundamental theoretical questions remain open:
(i) \emph{Finite-sample identifiability}: Under what conditions on local sample sizes $\{n_k\}_{k=1}^{K}$ and data heterogeneity is the global causal structure identifiable? How does federation affect the well-known identifiability-up-to-MEC result of~\cite{spirtes2000causation,chickering2002optimal}?
(ii) \emph{Minimax rates}: What are the minimax optimal rates for federated causal discovery under various heterogeneity models? That is, what is $\inf_{\hat{\mathcal{G}}} \sup_{\mathcal{G}^* \in \mathfrak{G}} \mathbb{E}[\Delta(\hat{\mathcal{G}}, \mathcal{G}^*)]$ over the class $\mathfrak{G}$ of DAGs under federated data partition constraints?
(iii) \emph{Semiparametric efficiency for FCI}: Extending the semiparametric efficiency bound theory~\cite{tsiatis2006semiparametric} to arbitrary federated topologies and heterogeneity patterns, determining whether and when federated estimators can achieve the same efficiency as centralized ones.
(iv) \emph{Consistency under model misspecification}: How robust are federated causal methods when structural or parametric assumptions are violated at a subset of sites?

\subsection{Vertical and Hybrid Federation}

The literature is overwhelmingly focused on horizontal FL.
Vertical and hybrid settings, which are arguably more common in practice (different institutions collect different variables about overlapping entities), remain underexplored.

\textbf{Open problems.}
In vertical FL, the fundamental challenge is that causal relationships \emph{between} variables held by different clients cannot be assessed locally.
Formally, if client $k$ holds variables $\mathbf{V}_k$ and $\mathbf{V}_j \cap \mathbf{V}_k = \emptyset$ for $j \neq k$ (with shared entity identifiers), then any edge $X_i \to X_j$ where $X_i \in \mathbf{V}_k$ and $X_j \in \mathbf{V}_\ell$ ($k \neq \ell$) requires cross-client information exchange.
Designing protocols that can test $X_i \perp\!\!\!\perp X_j \mid \mathbf{S}$ when $X_i$, $X_j$, and elements of $\mathbf{S}$ are distributed across clients, while preserving privacy and minimizing communication, is an open challenge.
FedISHC~\cite{chen2026horizontal} and fedCI-IOD~\cite{hahn2026federated} are initial steps, but principled solutions for arbitrary overlapping variable sets, handling latent confounders in the vertical setting~\cite{colombo2012learning,zhang2008completeness}, and scalable protocols for many-party vertical FL remain lacking.

\subsection{Bridging Discovery and Inference}

Currently, FCD and FCI are largely studied in isolation, despite their natural complementarity.

\textbf{Open problems.}
End-to-end federated pipelines that seamlessly transition from structure learning to effect estimation -- sharing structural information to improve downstream inference while maintaining privacy -- represent a promising but largely unexplored direction.
Key challenges include: (i) propagating uncertainty from the estimated DAG $\hat{\mathcal{G}}$ into the downstream effect estimate $\hat{\tau}$, since ignoring structural uncertainty can lead to over-confident causal conclusions~\cite{maathuis2009estimating}; (ii) designing joint optimization objectives that simultaneously learn structure and estimate effects, potentially achieving better statistical efficiency than a two-stage approach; and (iii) developing federated sensitivity analysis methods~\cite{ding2016sensitivity} that quantify the robustness of causal conclusions to unmeasured confounding across sites.
FedECE~\cite{zhao2025fedece} provides an initial realization, but much more work is needed.

\subsection{Causal Reasoning Under Complex Settings}

Several practically important causal settings lack federated solutions:

\textbf{(i) Latent confounders and selection bias.}
Most FCD methods assume causal sufficiency (no unmeasured confounders), yet in practice, unmeasured confounders are ubiquitous.
Extending federated methods to learn partial ancestral graphs (PAGs)~\cite{zhang2008completeness,colombo2012learning}, which represent equivalence classes under latent confounding, is an important direction.

\textbf{(ii) Time-varying treatments.}
In longitudinal settings, treatments are applied sequentially over time, with time-varying confounding.
Extending federated methods to handle dynamic treatment regimes~\cite{murphy2003optimal}, including the federated estimation of marginal structural models (MSMs) with time-varying inverse probability weights $\prod_{t=1}^{T} w_t(\bar{A}_t, \bar{L}_t)$, is a challenging open problem.

\textbf{(iii) Mediation analysis.}
Decomposing the total causal effect into direct and indirect (mediated) pathways requires estimating natural direct and indirect effects~\cite{vanderweele2015explanation}.
In federated settings, the mediator $M$ and outcome $Y$ may reside at different sites, complicating cross-site identification.

\textbf{(iv) Counterfactual reasoning.}
Moving beyond interventional queries ($P(Y \mid do(X))$) to counterfactual queries ($P(Y_{X=x'} \mid X=x, Y=y)$)~\cite{pearl2009causality} in federated settings requires reasoning about individual-level potential outcomes across sites, which poses unique privacy and statistical challenges.

\textbf{(v) Causal reinforcement learning.}
Integrating causal reasoning into federated reinforcement learning~\cite{zhang2020designing} could enable distributed agents to learn optimal policies while accounting for confounding in observational data.

\subsection{Large Language Models and Foundation Models}

The rapid development of large language models (LLMs) and foundation models opens new avenues for federated causal reasoning~\cite{kiciman2024causal,ban2023query,jin2023cladder}.

\textbf{Open problems.}
LLMs can serve as sources of prior causal knowledge (\eg, suggesting plausible causal directions between variables based on domain knowledge encoded in their training data~\cite{kiciman2024causal}), which could constrain the search space in federated causal discovery and reduce communication overhead.
Conversely, federated causal methods could improve the reliability and factual grounding of LLM outputs by enabling privacy-preserving validation of LLM-generated causal claims against distributed observational data.
Key challenges include: (i) formalizing how to integrate LLM-based causal priors as informative constraints in federated structure learning algorithms; (ii) developing federated fine-tuning pipelines that improve LLM causal reasoning capabilities without centralizing domain-specific data; and (iii) ensuring that LLM-assisted causal conclusions maintain formal statistical guarantees despite the informal nature of LLM priors.

\subsection{Benchmarks and Reproducibility}

The field lacks standardized benchmarks for comparing FCD and FCI methods.
Most papers use different synthetic data generators, real datasets, and discrete BN datasets, making cross-paper comparisons difficult.

\textbf{Open problems.}
Establishing community benchmarks with realistic federated settings is essential.
Such benchmarks should encompass: (i) data heterogeneity models with controlled degrees of distributional shift; (ii) varying numbers of clients, sample sizes, and variable dimensions; (iii) standardized missing data scenarios with specified missingness mechanisms (MCAR, MAR, MNAR)~\cite{mohan2013graphical}; (iv) known ground-truth causal structures and treatment effects for rigorous evaluation; and (v) realistic privacy constraints and communication budgets.
Metrics should go beyond structural accuracy (SHD, F1 score) to include causal effect estimation error ($|\hat{\tau} - \tau|$, RMSE of CATE), privacy leakage measures, and communication cost.

\subsection{Scalability to Large Variable Sets}

While FedCSL-HD~\cite{ding2025federated} and FedCSL~\cite{guo2024fedcsl} begin to address high dimensionality, scaling FCD to hundreds or thousands of variables in federated settings remains challenging, as the DAG space grows super-exponentially with $d$ (the number of possible DAGs over $d$ nodes grows as $O(d! \cdot 2^{\binom{d}{2}})$~\cite{robinson2006counting}).

\textbf{Open problems.}
Divide-and-conquer strategies~\cite{ding2025federated}, federated variable selection, and causal feature screening (\eg, identifying the Markov blanket of the target variable~\cite{tsamardinos2003time,yu2020causality}) deserve further attention.
Additionally, leveraging distributed computing paradigms beyond the standard client-server architecture (such as hierarchical or peer-to-peer federation~\cite{gupta2023travellingfl}) could enable more scalable federated causal discovery.

\section{Conclusion}\label{sec:conclusion}

This survey has provided a comprehensive and focused review of federated causal discovery and inference, offering what we believe to be the most in-depth and systematically organized treatment of both subfields to date.

We organized the FCD literature along three taxonomic axes (methodology: constraint-based, score-based, continuous optimization, and hybrid; federation topology: horizontal, vertical, and hybrid; and structural scope: global vs.\ local) and the FCI literature by target estimand (ATE vs.\ ITE/CATE) and estimation strategy (IPW, AIPW, matching, Bayesian, deep generative, DML, IV, and survival analysis methods).
Our analysis reveals several important trends and gaps:
(i)~the dominance of horizontal FL settings, with vertical and hybrid federation remaining significantly underexplored despite their practical prevalence;
(ii)~the growing attention to data heterogeneity and non-identical variable sets, moving beyond the initial assumption of homogeneous data-generating processes;
(iii)~the scarcity of formal privacy guarantees, with the majority of methods relying on the implicit assumption that not sharing raw data suffices;
(iv)~the underexplored connection between discovery and inference, despite their natural complementarity as stages of a unified causal reasoning pipeline;
and (v)~the need for standardized benchmarks, theoretical foundations (especially minimax rates and semiparametric efficiency bounds under federated constraints), and scalable solutions for high-dimensional settings.

Compared with the broader Causal-FL perspective surveyed in~\cite{deng2026survey}, which encompasses how causality enhances FL systems in dimensions such as interpretability, generalizability, and robustness, our work provides a dedicated, algorithm-level treatment of the core problems of federated structure learning and effect estimation.
We believe these two complementary perspectives together paint a comprehensive picture of the rapidly evolving landscape at the intersection of causality and federated learning.
We hope this survey serves as both a reference for current work and a roadmap for future research, ultimately helping to realize the full potential of federated causal reasoning in domains ranging from healthcare to manufacturing, economics, and beyond.

\Acknowledgements{This work was supported by the National Natural Science Foundation of China (Grant Nos. 62506174, 62376087), the Basic Research Program of Jiangsu (Grant No. BK20250656), and the Natural Science Research Start-up Foundation of Recruiting Talents of Nanjing University of Posts and Telecommunications (Grant No. NY225023).}



\bibliographystyle{scis}
\bibliography{reference}



\end{document}